%% file: root.tex
\documentclass[letterpaper, 10 pt, conference]{ieeeconf}  
\IEEEoverridecommandlockouts
\usepackage{amsmath,amsfonts}
\usepackage{algorithm}
\usepackage[noend]{algpseudocode}
\usepackage{array}
\usepackage{balance}
\usepackage[font=scriptsize,labelfont=bf]{caption}
\usepackage[nomath]{cellspace}
\usepackage{duckuments}
\usepackage{float}
\usepackage{graphicx}
\usepackage{mathtools}
\usepackage{multicol}
\usepackage{multirow}
\usepackage[numbers,sort&compress]{natbib}
\usepackage{stfloats}
\usepackage[font=scriptsize,labelfont=bf]{subcaption}
\usepackage{titlesec}
\usepackage{textcomp}
\usepackage{url}
\usepackage{verbatim}
\usepackage[dvipsnames]{xcolor}

\usepackage[bookmarks=true]{hyperref}
\usepackage{cleveref}
\hypersetup{
    colorlinks=true,
    linkcolor=blue,
    filecolor=magenta,      
    urlcolor=cyan,
    pdfpagemode=FullScreen,
}

\AtBeginEnvironment{tabular}{\scriptsize}
\IEEEaftertitletext{\vspace{-1.4\baselineskip}}
\titlespacing\subsection{0pt}{6pt plus 2pt minus 2pt}{2pt plus 2pt minus 2pt}
\titleformat{\subsubsection}[runin]
   {\itshape}
   {\thesubsubsectiondis}
   {0.5em}
   {}
   [:]
\titlespacing\subsubsection{\parindent}{0pt}{0.5em}

\title{\vspace{-5pt}\LARGE \bf COSMO-Bench: A Benchmark for Collaborative SLAM Optimization \vspace{-16pt} 
\author{Daniel McGann, Easton R. Potokar, and Michael Kaess}
}

\input{helpers}

\begin{document}

\twocolumn[{%
    \renewcommand\twocolumn[1][]{#1}
	\maketitle
	\thispagestyle{empty}
    \vspace{-1.15cm}
    \begin{center}
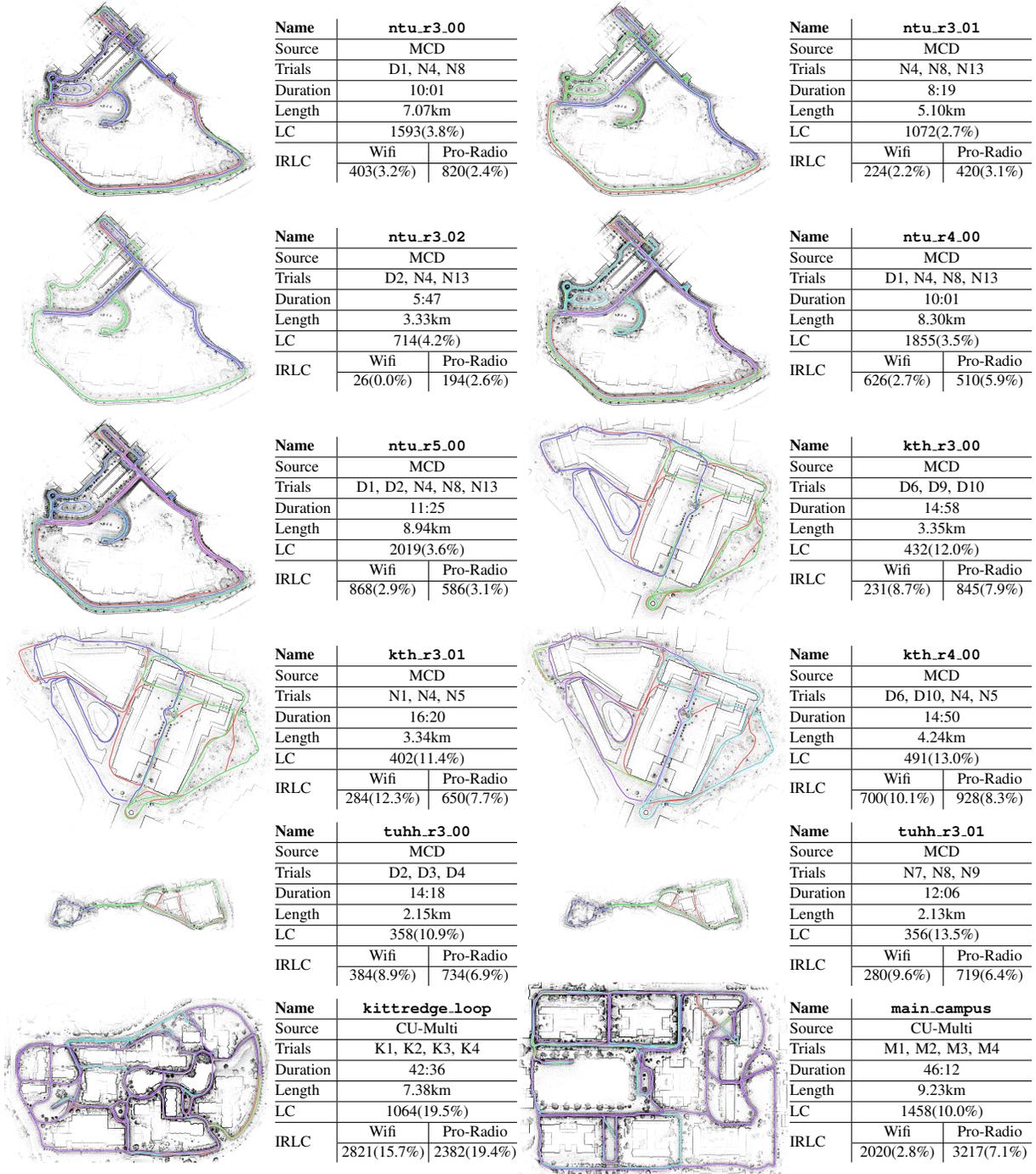

        \centering
        \include{figs/dataset_figures/new_dataset_figure_grid}
        \vspace{-0.6cm}
        \captionof{figure}{The COSMO-Bench datasets. For each sequence, we plot the reference solution and enumerate metadata including -- The sequence name, source data, component trials, total duration (MM:SS), total distance traveled, and the number (\#) of measurements plus outlier rate (\%) for both intra-robot (LC) and inter-robot (IRLC) loop-closures (\#, \%). For each sequence, we generate a dataset using both the Wi-Fi and Pro-Radio communication models for a total of 24 datasets. Component trial names are shortened for brevity -- "D" for "Day" and "N" for "Night" for the MCD data and "K" for "Kittredge Loop" and "M" for "Main Campus" for the CU-Multi data. Note: Plots are not to scale.}
        \label{fig:datasets}
        \vspace{-12pt}
    \end{center}
}]

\thispagestyle{empty}
\pagestyle{empty}

\begin{abstract}
Recent years have seen a focus on research into distributed optimization algorithms for multi-robot Collaborative Simultaneous Localization and Mapping (C-SLAM). Research in this domain, however, is made difficult by a lack of standard benchmark datasets. Such datasets have been used to great effect in the field of single-robot SLAM, and researchers focused on multi-robot problems would benefit greatly from dedicated benchmark datasets. To address this gap, we design and release the \textit{C}ollaborative \textit{O}pen-\textit{S}ource \textit{M}ulti-robot \textit{O}ptimization \textit{Bench}mark (COSMO-Bench) -- a suite of 24 datasets derived from a baseline C-SLAM front-end and real-world LiDAR data. Data DOI: ~\href{https://doi.org/10.1184/R1/29652158}{\texttt{10.1184/R1/29652158}}.
\vspace{-4pt}%
\blfootnote{\scriptsize\hspace{-8.5pt}This work was partially supported by NASA award 80NSSC24CA020 and the NSF Graduate Research Fellowship Program.}
\blfootnote{\scriptsize\hspace{-8.5pt}The authors are with the Robotics Institute, Carnegie Mellon University, Pittsburgh, PA, USA. \texttt{\{danmcgann, potokar, kaess\}@cmu.edu}}
\end{abstract}

\input{sections/0-intro}

\input{sections/1-preliminaries}
\input{sections/2-related_work}
\input{sections/3-methodology}

\input{sections/4-cosmobench-datasets}
\input{sections/5-neubla-dataset-adaption}

\input{sections/6-conclusion}

\bibliographystyle{ieeetr} 
\footnotesize
\bibliography{refs}

\end{document}

%% file: helpers.tex
\usepackage[dvipsnames]{xcolor}
\usepackage{amsfonts}

\usepackage{amsthm}

\newcommand\blfootnote[1]{
    \begingroup
    \renewcommand\thefootnote{}\footnote{#1}
    \addtocounter{footnote}{-1}
    \endgroup
}
\newtheoremstyle{dense}
  {3pt} 
  {3pt} 
  {\itshape} 
  {} 
  {\bfseries} 
  {:} 
  {.5em} 
  {} 
\theoremstyle{dense}

\newtheorem{remark}{Remark}
\AtBeginEnvironment{remark}{\small}
\AtBeginEnvironment{quote}{\par\singlespacing\small}

\definecolor{tab:blue}{RGB}{31,119,180}
\definecolor{tab:orange}{RGB}{255,127,14}
\definecolor{tab:green}{RGB}{44,160,44}
\definecolor{tab:red}{RGB}{214,39,40}
\definecolor{tab:purple}{RGB}{148,103,189}

\definecolor{sns:blue}{rgb}{0.00392156862745098, 0.45098039215686275, 0.6980392156862745}
\definecolor{sns:orange}{rgb}{0.8705882352941177, 0.5607843137254902, 0.0196078431372549}
\definecolor{sns:green}{rgb}{0.00784313725490196, 0.6196078431372549, 0.45098039215686275}
\definecolor{sns:red}{rgb}{0.8352941176470589, 0.3686274509803922, 0.0}
\definecolor{sns:purple}{rgb}{0.8, 0.47058823529411764, 0.7372549019607844}
\definecolor{sns:brown}{rgb}{0.792156862745098, 0.5686274509803921, 0.3803921568627451}
\definecolor{sns:pink}{rgb}{0.984313725490196, 0.6862745098039216, 0.8941176470588236}
\definecolor{sns:gray}{rgb}{0.5803921568627451, 0.5803921568627451, 0.5803921568627451}
\definecolor{sns:yellow}{rgb}{0.9254901960784314, 0.8823529411764706, 0.2}
\definecolor{sns:light_blue}{rgb}{0.33725490196078434, 0.7058823529411765, 0.9137254901960784}

\newcommand{\logmap}[1]{\mathrm{Log}\left(#1\right)}

\newcommand{\Bc}{\mathcal{B}}
\newcommand{\Cc}{\mathcal{C}}

\newcommand{\Ic}{\mathcal{I}}

\newcommand{\Nc}{\mathcal{N}}

\newcommand{\Tc}{\mathcal{T}}



%% file: figs/dataset_figures/new_dataset_figure_grid.tex
\setlength{\fboxsep}{0pt}

\newcommand\infotable[9]{
\begin{minipage}[b][#9][c]{1.5in}
\centering
\begin{tabular}{@{}p{0.35in}|cc@{}}
    \textbf{Name}           & \multicolumn{2}{p{1.05in}}{\centering\texttt{\textbf{#1}}}  \\ \hline
    Source                  & \multicolumn{2}{c}{#2}                    \\ \hline
    Trials                  & \multicolumn{2}{c}{#3}                    \\ \hline
    Duration                & \multicolumn{2}{c}{#4}                    \\ \hline
    Length                  & \multicolumn{2}{c}{#5}                    \\ \hline
    LC                      & \multicolumn{2}{c}{#6}                    \\ \hline
    \multirow{2}{*}{IRLC}   & \multicolumn{1}{p{0.5025in}|}{\centering{Wifi}}     & Pro-Radio \\ \cline{2-3} 
                            & \multicolumn{1}{p{0.5025in}|}{\centering{#7}}       & #8        \\ 
\end{tabular}\end{minipage}}

\renewcommand{\arraystretch}{1.1}
\setlength{\tabcolsep}{2pt} 

\begin{tabular}[c]{@{}c@{}c@{}c@{}c@{}}
\includegraphics[height=1.25in]{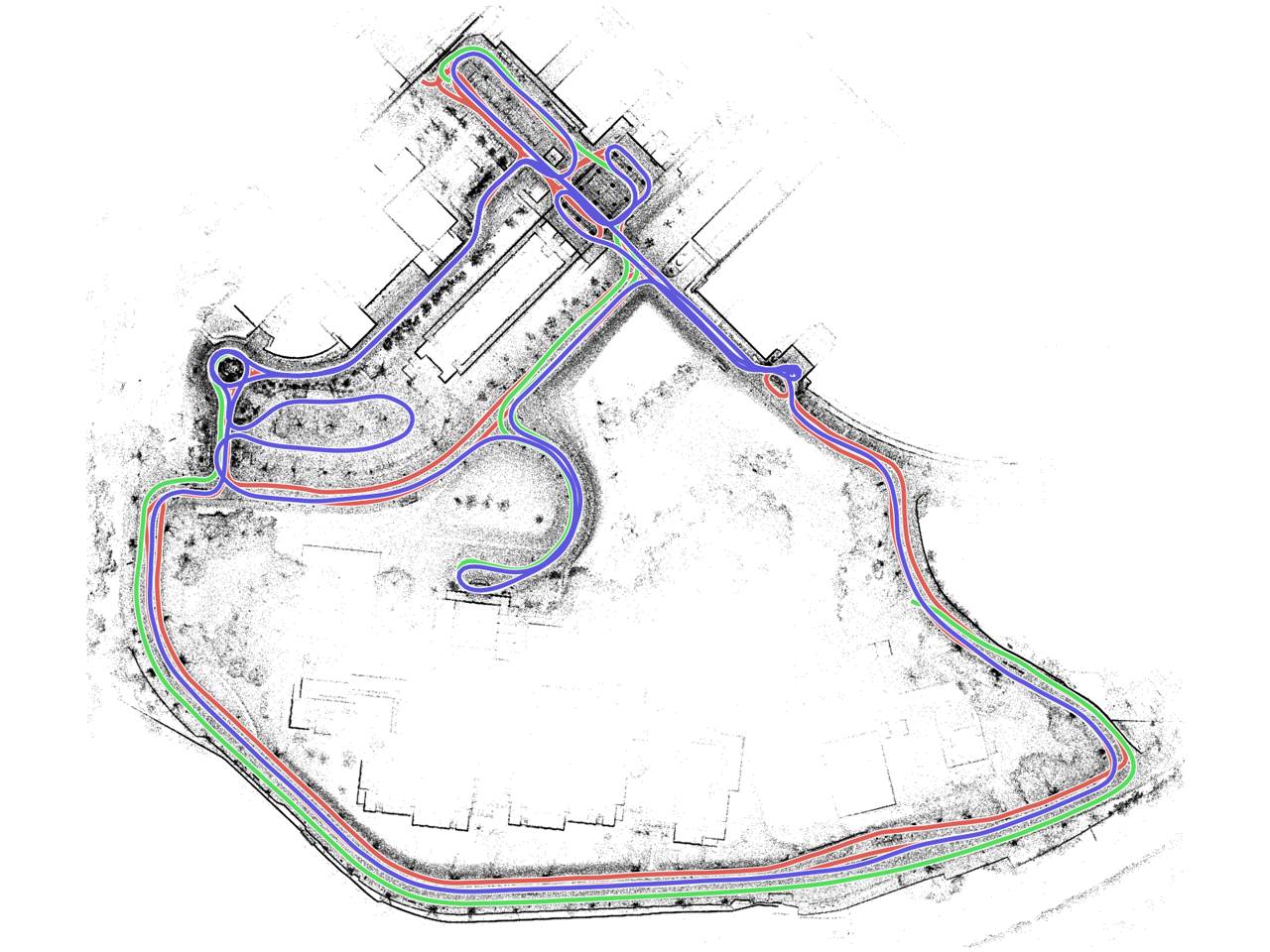} &
\infotable{ntu\_r3\_00}{MCD}{D1, N4, N8}{10:01}{7.07km}{1593(3.8\%)}{403(3.2\%)}{820(2.4\%)}{1.25in} &

\includegraphics[height=1.25in]{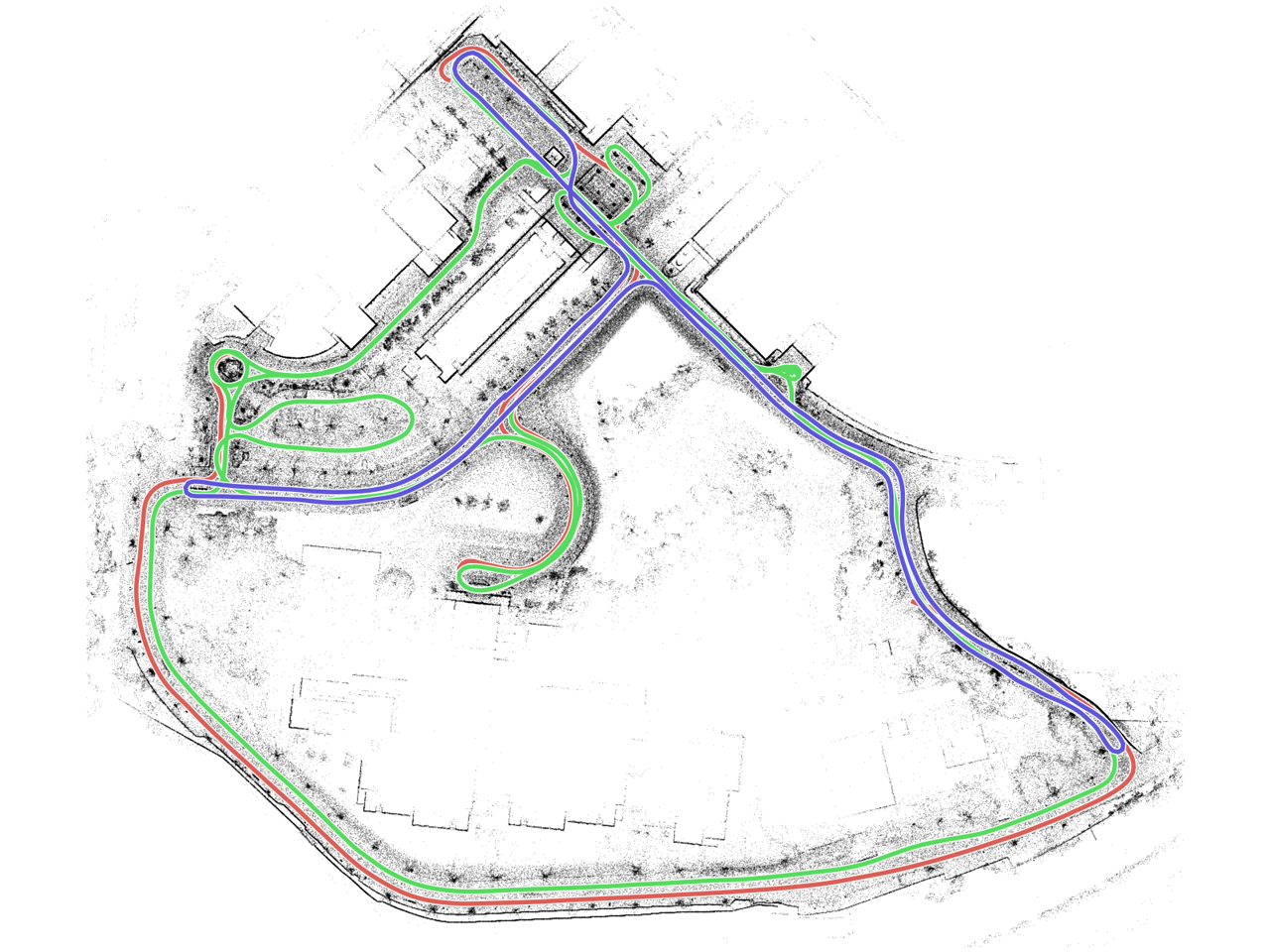} &
\infotable{ntu\_r3\_01}{MCD}{N4, N8, N13}{8:19}{5.10km}{1072(2.7\%)}{224(2.2\%)}{420(3.1\%)}{1.25in} \\

\includegraphics[height=1.25in]{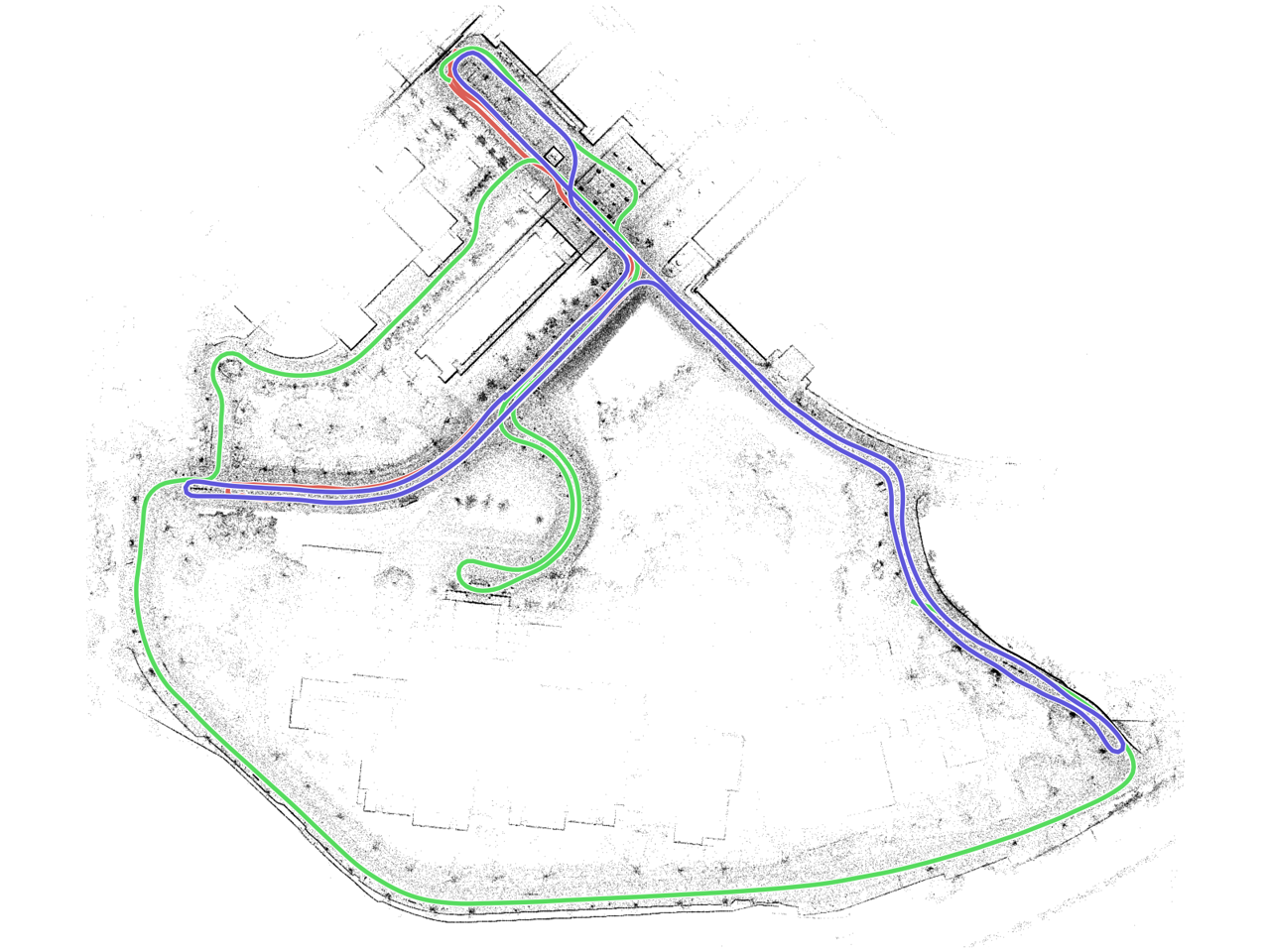} &
\infotable{ntu\_r3\_02}{MCD}{D2, N4, N13}{5:47}{3.33km}{714(4.2\%)}{26(0.0\%)}{194(2.6\%)}{1.25in} &

\includegraphics[height=1.25in]{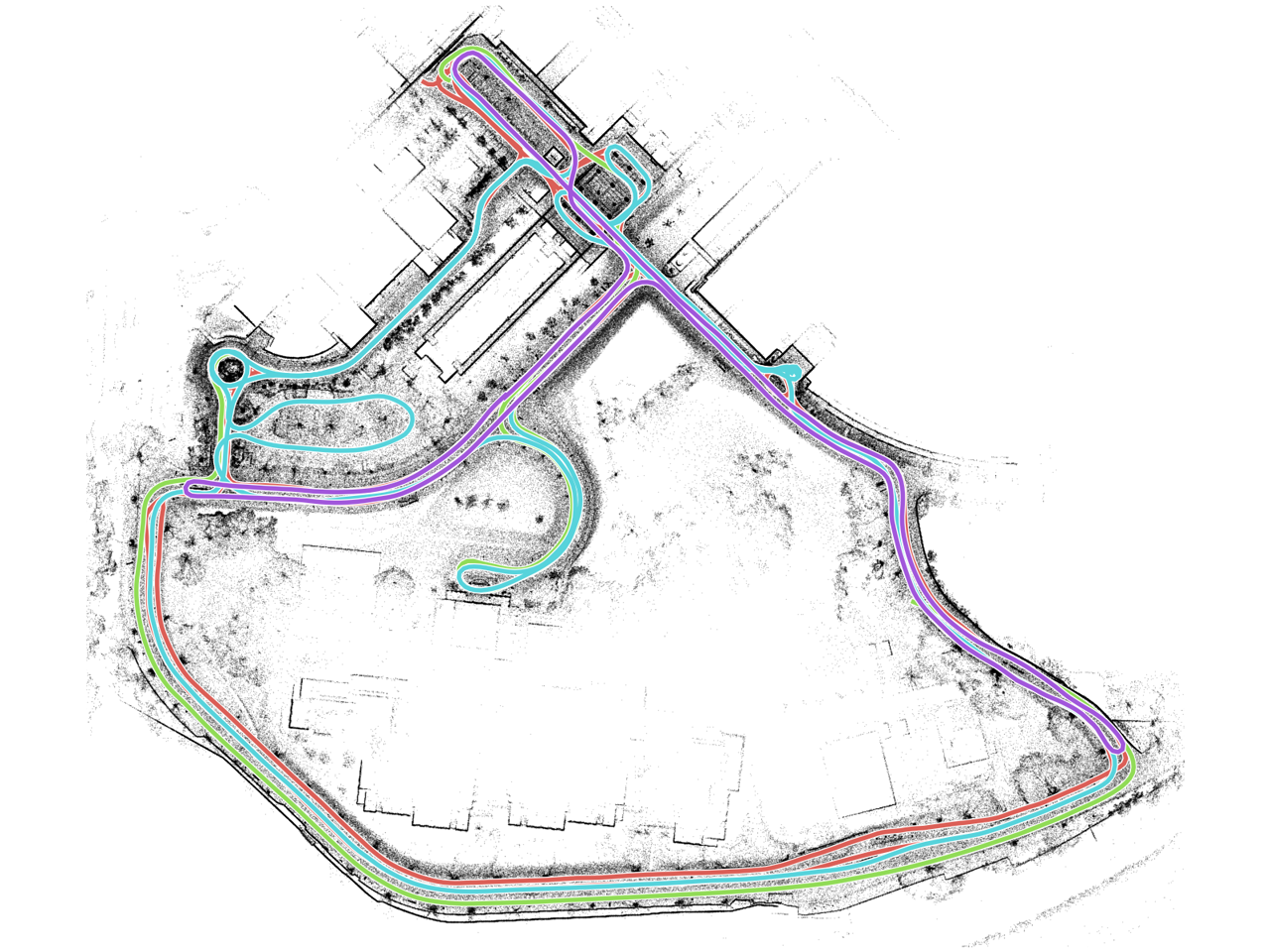} &
\infotable{ntu\_r4\_00}{MCD}{D1, N4, N8, N13}{10:01}{8.30km}{1855(3.5\%)}{626(2.7\%)}{510(5.9\%)}{1.25in} \\

\includegraphics[height=1.25in]{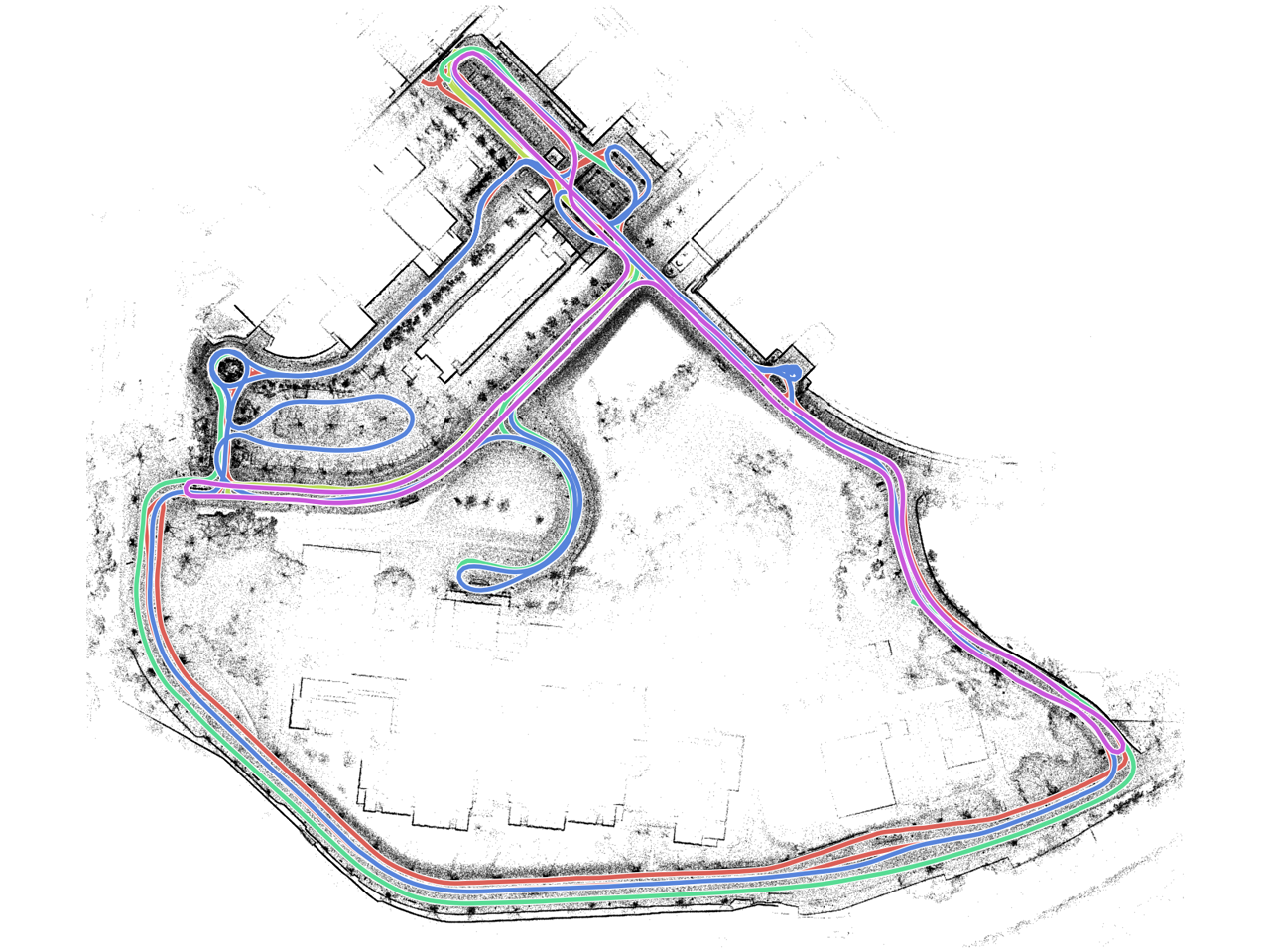} &
\infotable{ntu\_r5\_00}{MCD}{D1, D2, N4, N8, N13}{11:25}{8.94km}{2019(3.6\%)}{868(2.9\%)}{586(3.1\%)}{1.25in} &

\includegraphics[height=1.25in]{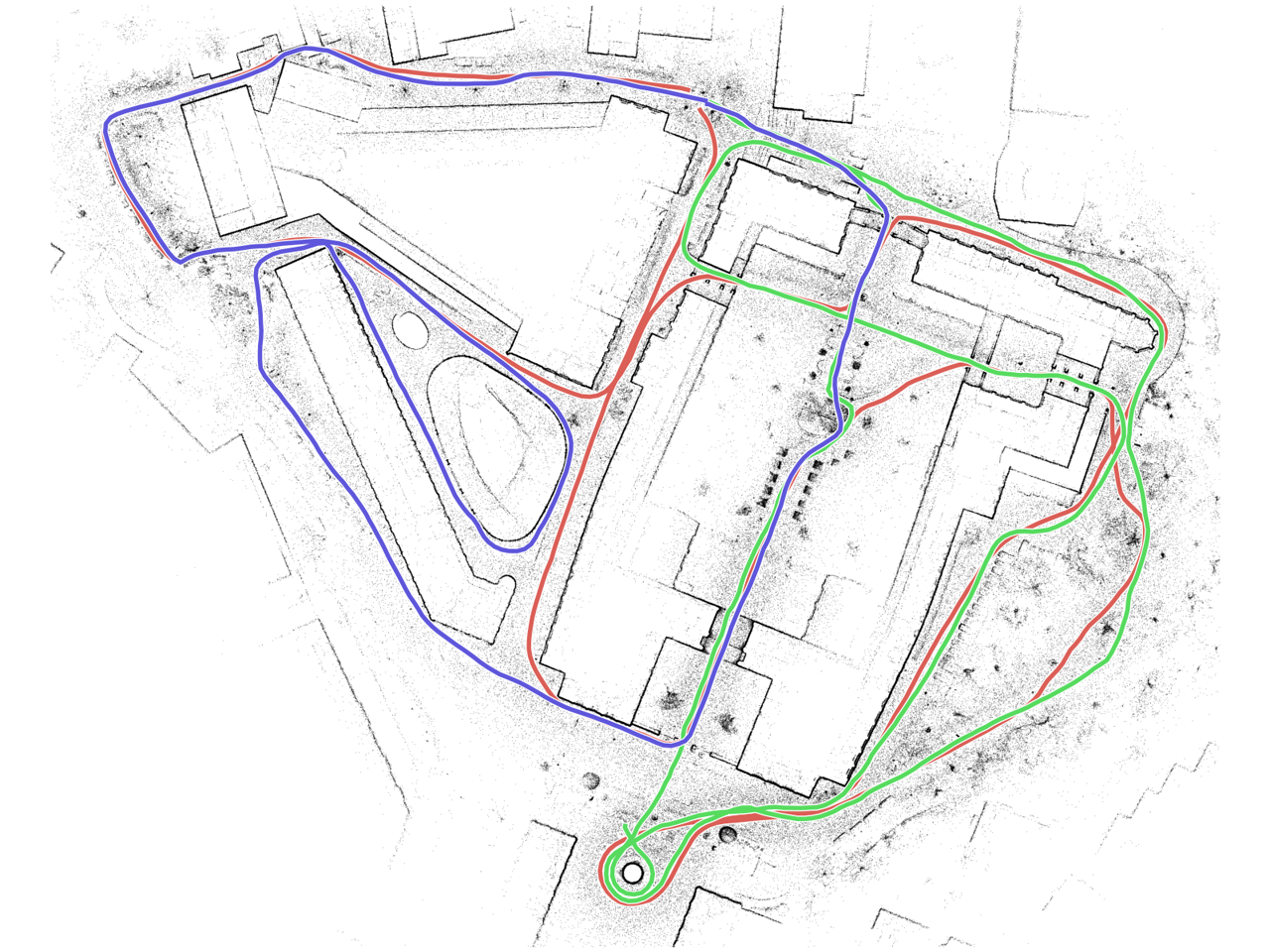} &
\infotable{kth\_r3\_00}{MCD}{D6, D9, D10}{14:58}{3.35km}{432(12.0\%)}{231(8.7\%)}{845(7.9\%)}{1.25in} \\

\includegraphics[height=1.25in]{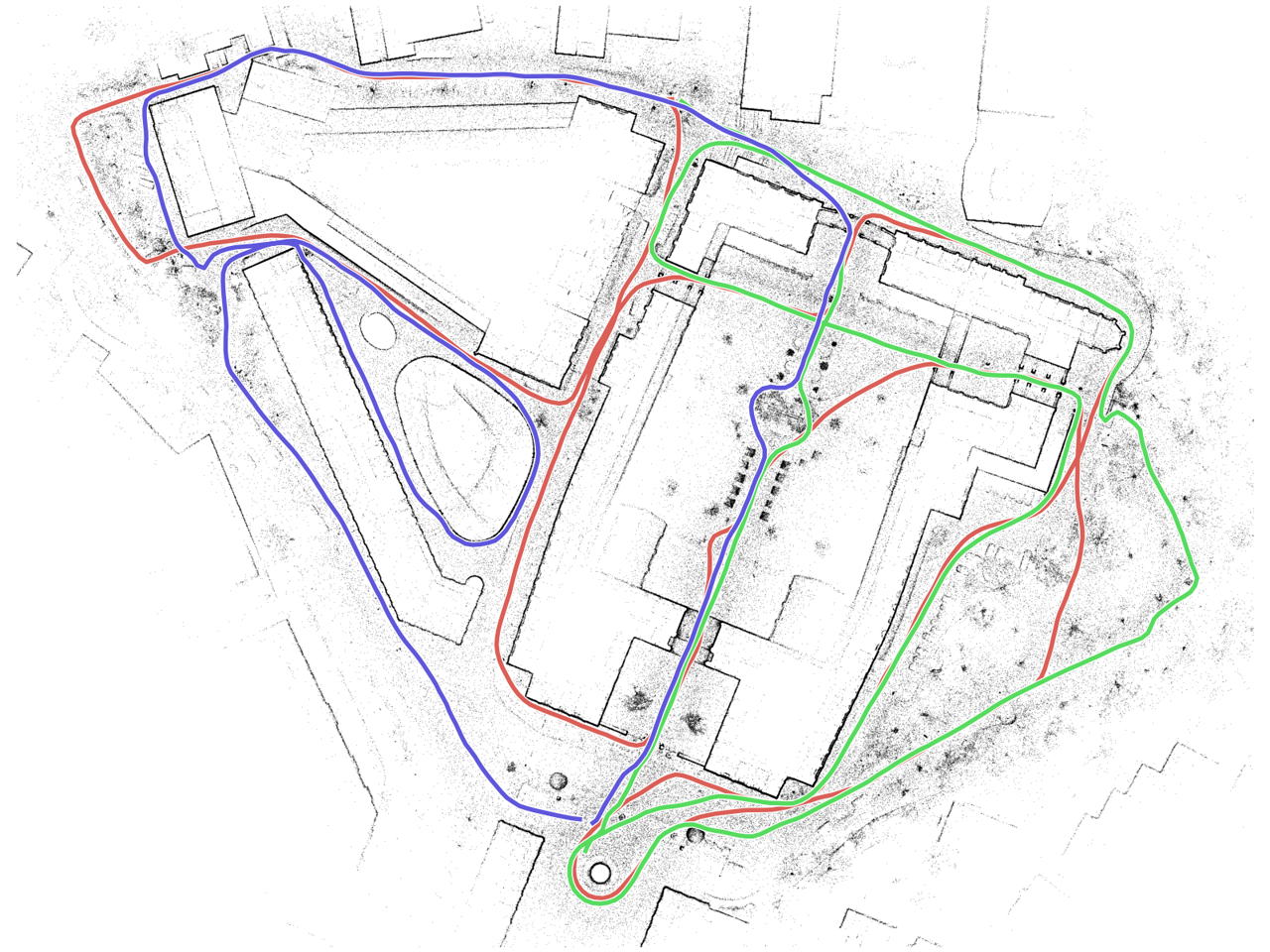} &
\infotable{kth\_r3\_01}{MCD}{N1, N4, N5}{16:20}{3.34km}{402(11.4\%)}{284(12.3\%)}{650(7.7\%)}{1.25in} &

\includegraphics[height=1.25in]{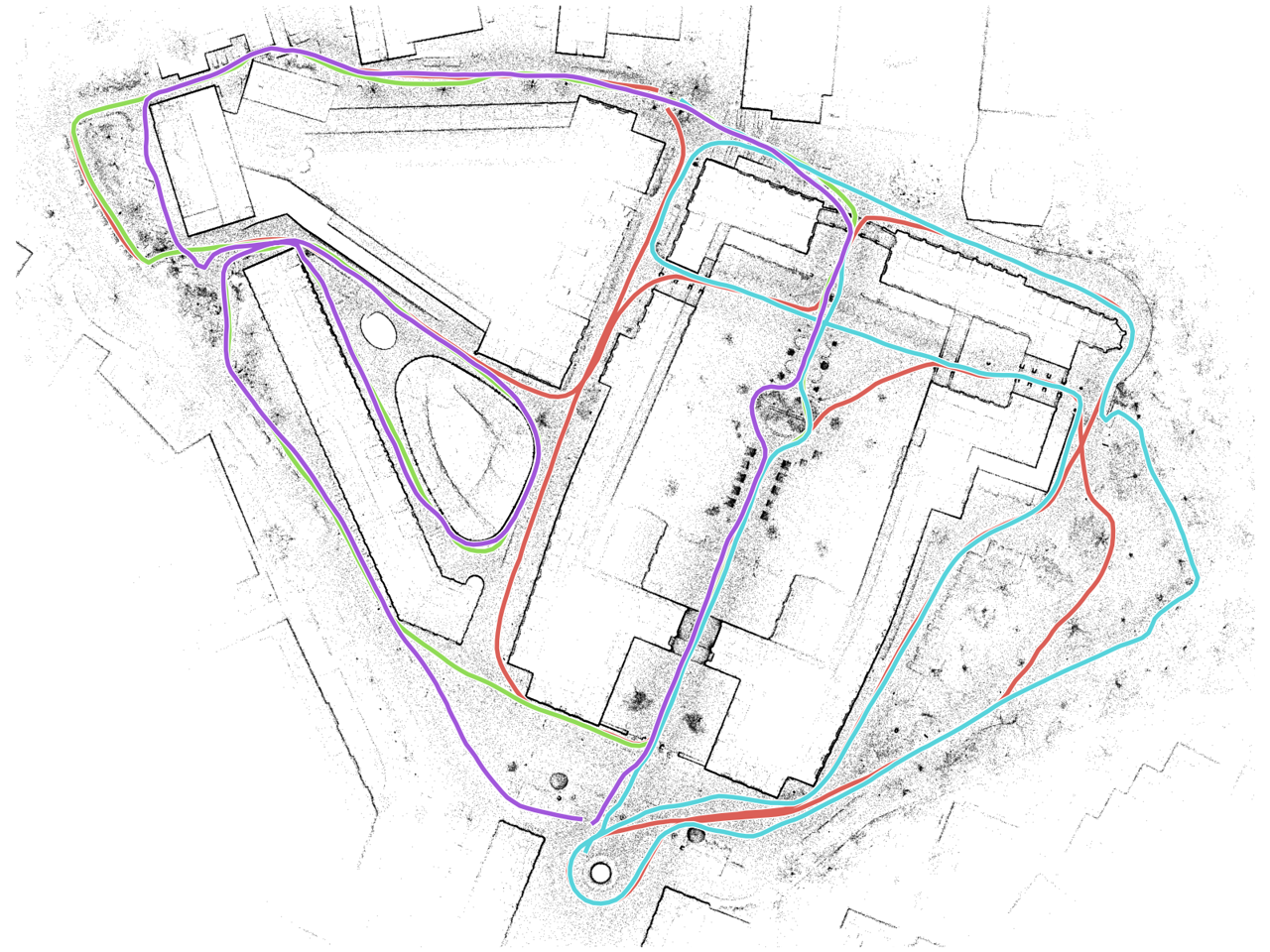} &
\infotable{kth\_r4\_00}{MCD}{D6, D10, N4, N5}{14:50}{4.24km}{491(13.0\%)}{700(10.1\%)}{928(8.3\%)}{1.25in} \\

\includegraphics[trim={0 0.2in 0 0.2in},clip,height=0.85in]{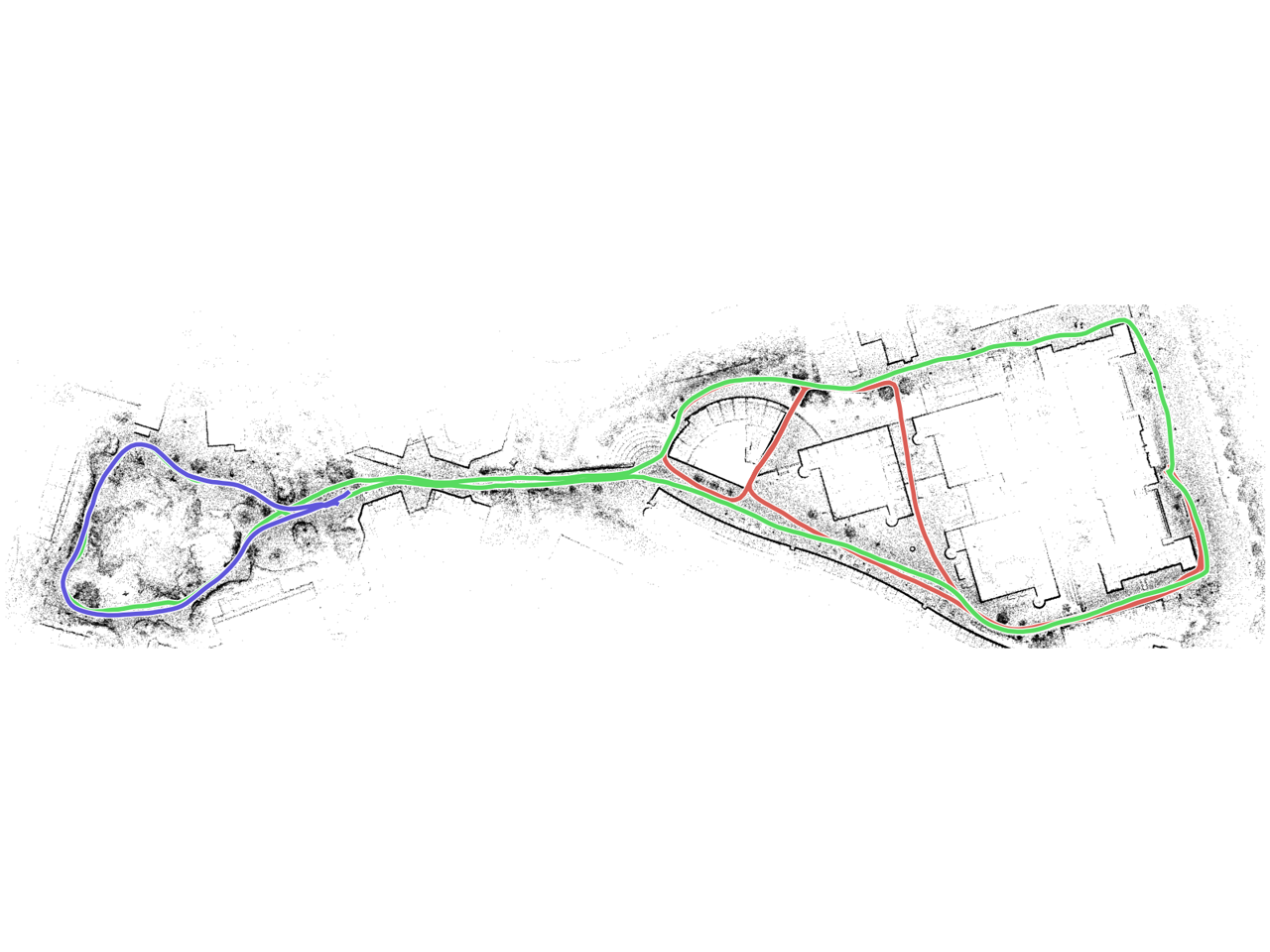} &
\infotable{tuhh\_r3\_00}{MCD}{D2, D3, D4}{14:18}{2.15km}{358(10.9\%)}{384(8.9\%)}{734(6.9\%)}{0.85in} &

\includegraphics[trim={0 0.2in 0 0.2in},clip,height=0.85in]{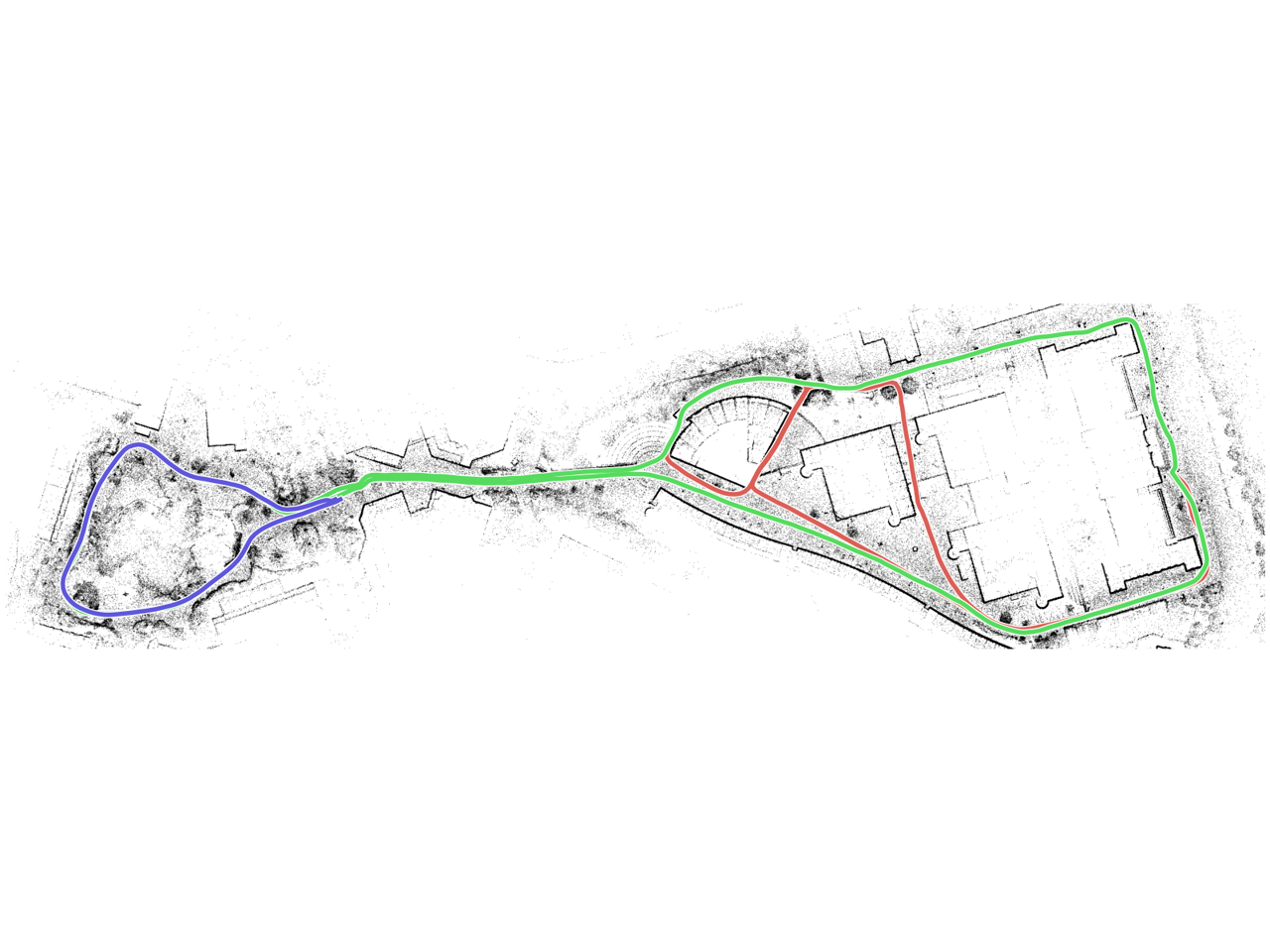} &
\infotable{tuhh\_r3\_01}{MCD}{N7, N8, N9}{12:06}{2.13km}{356(13.5\%)}{280(9.6\%)}{719(6.4\%)}{0.85in} \\

\includegraphics[height=1.25in]{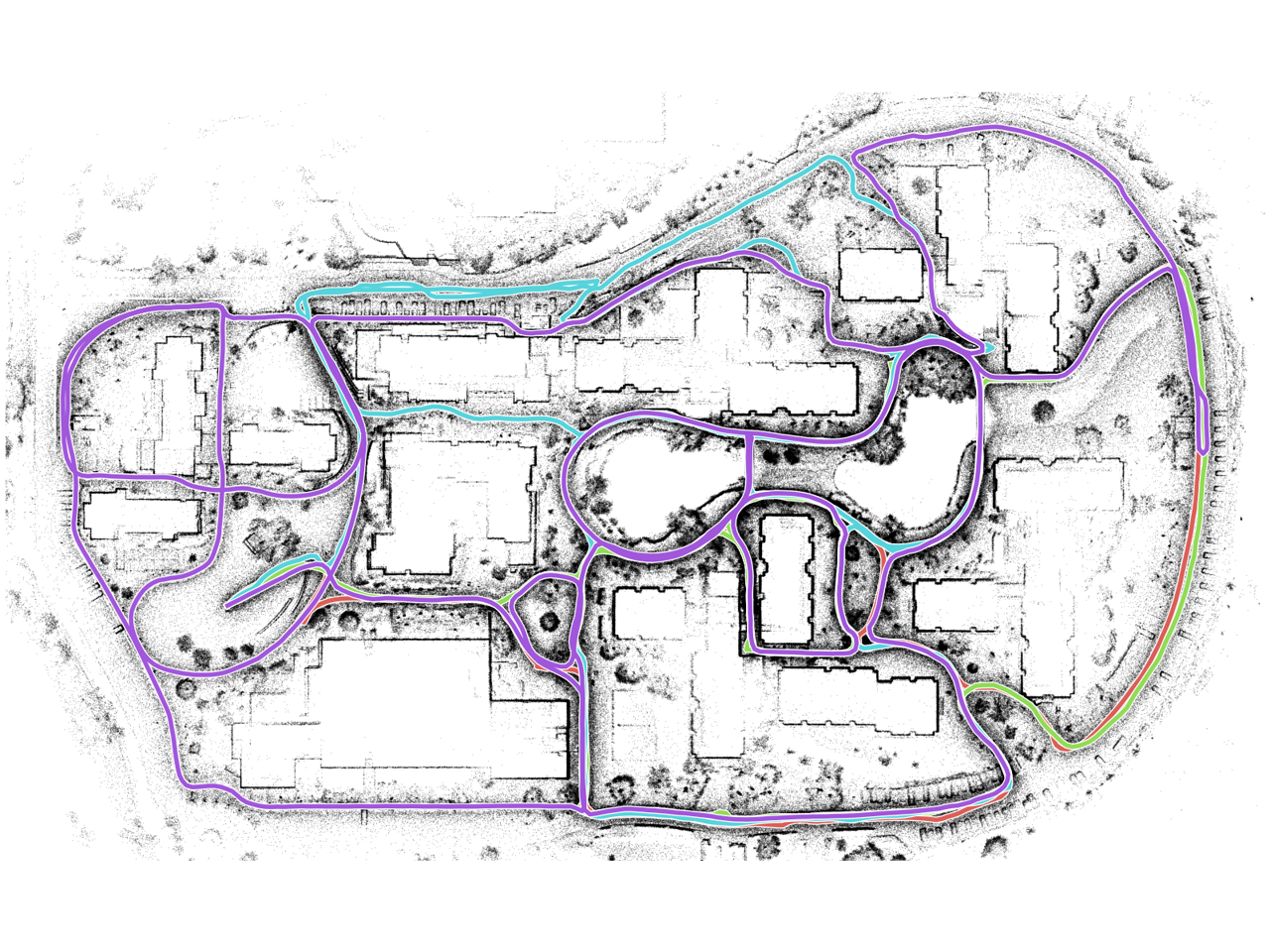} &
\infotable{kittredge\_loop}{CU-Multi}{K1, K2, K3, K4}{42:36}{7.38km}{1064(19.5\%)}{2821(15.7\%)}{2382(19.4\%)}{1.25in} &

\includegraphics[height=1.25in]{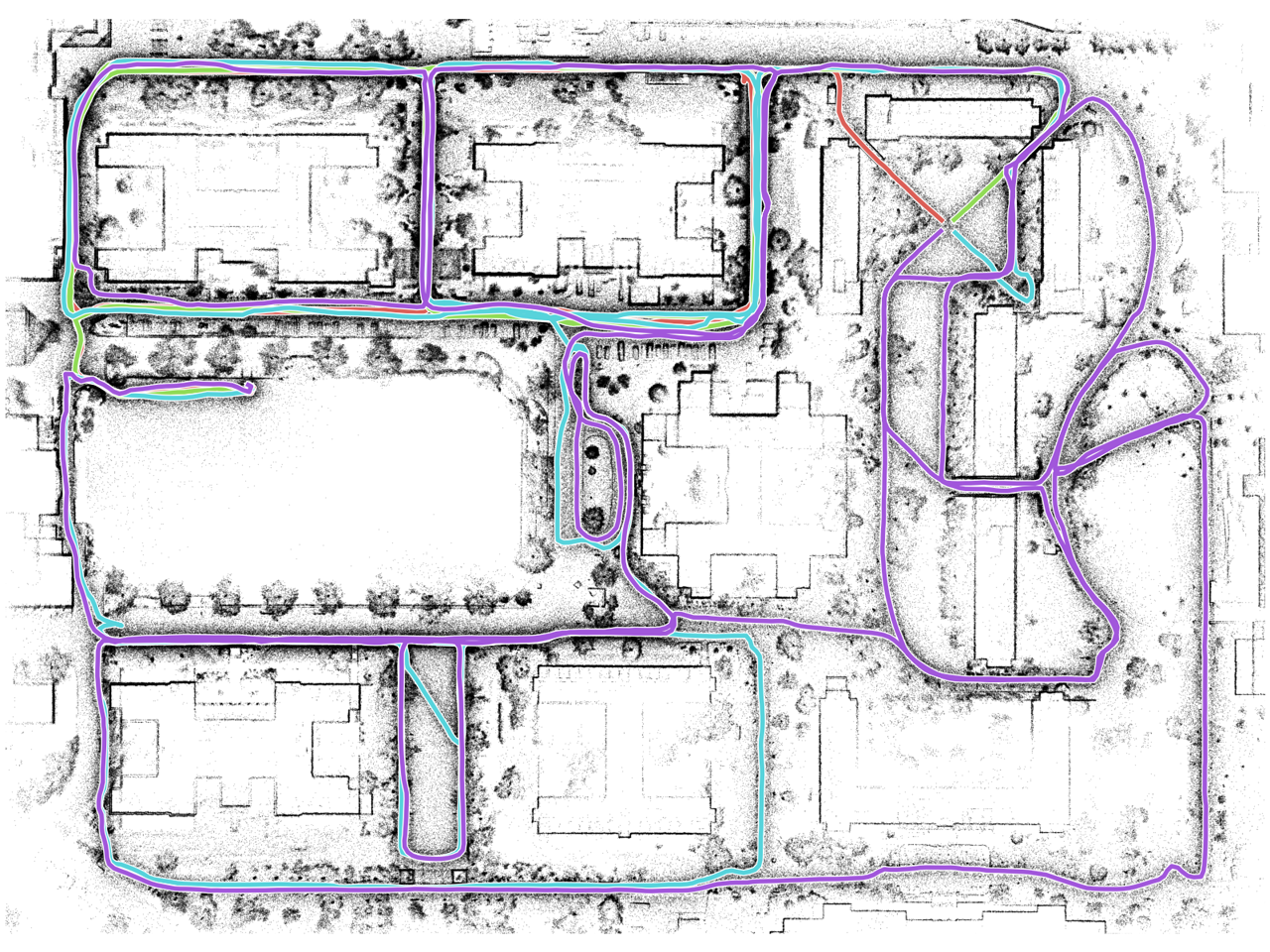} &
\infotable{main\_campus}{CU-Multi}{M1, M2, M3, M4}{46:12}{9.23km}{1458(10.0\%)}{2020(2.8\%)}{3217(7.1\%)}{1.25in} \\
\end{tabular}

%% file: sections/0-intro.tex
\section{Introduction}\label{sec:intro}

Recent years have seen a resurgence in research into distributed optimization algorithms for Multi-Robot Collaborative Simultaneous Localization and Mapping (C-SLAM)~\cite{lajoie_cslam_survey_2022}. These algorithms (often referred to as ``back-ends'') synthesize noisy and potentially incorrect measurements (produced by a ``front-end") into an optimal state estimate~\cite{slam_survey_leonard_2016}. However, a lack of standard benchmark datasets for distributed C-SLAM has hindered research into novel back-end methods.

In the single-robot SLAM domain, benchmark datasets (e.g. M3500~\cite{fast_olson_2006}, Intel~\cite{slam_eval_burgard_2009}, MIT-Killian~\cite{slam_eval_burgard_2009}, CSAIL~\cite{slam_eval_burgard_2009}, Garage~\cite{g2o_kummerle_2011}, etc.), which consist of geometric measurements output by a SLAM front-end, are widely used to develop and test novel algorithms~\cite{sesync_rosen_2019, imcbpgo_chen_2024, supernova_kim_2025}. Their use reduces the barrier of entry to try new ideas, provides a means by which researchers can reproduce results, enables relative performance comparisons across works, and broadly aids the community's mission to advance the state-of-the-art.

Researchers would benefit greatly from benchmark datasets for distributed C-SLAM back-ends. The need for such datasets is exemplified by recent work's reliance on single-robot benchmarks that have been partitioned to, unrealistically, simulate multi-robot scenarios~\cite{admmslam_choudhary_2015, geod_cristofalo_2020, dc2pgo_tian_2021, asapp_tian_2020, idfgo_matsuka_2023, maj_min_journal_fan_2024, mesa_mcgann_2024}.

In this work we develop the \textbf{C}ollaborative \textbf{O}pen-\textbf{S}ource \textbf{M}ulti-robot \textbf{O}ptimization \textbf{Bench}mark (COSMO-Bench)\footnote{Hosted in perpetuity at 
~\href{https://doi.org/10.1184/R1/29652158}{\texttt{doi.org/10.1184/R1/29652158}} and accessible with a user-friendly interface at \href{https://cosmobench.com}{\texttt{cosmobench.com}}.} -- a suite of 24 benchmark datasets for distributed C-SLAM back-end evaluation (Fig.~\ref{fig:datasets}). We work to make COSMO-Bench as realistic as possible by deriving the datasets from real-world sensor data, a baseline LiDAR-based C-SLAM front-end, and a communication model developed from actual multi-robot communication data.

%% file: sections/1-preliminaries.tex
\section{Benchmark Requirements}\label{sec:benchmark_reqs}

The most useful distributed C-SLAM benchmark datasets will be those that most accurately reflect the data that is observed by a multi-robot team in a real-world deployment. However, we additionally need benchmark datasets that permit accurate and insightful evaluation of C-SLAM back-end performance. In this section we outline the traits of a dataset that achieves these goals. These traits, in turn, act as the requirements for our proposed benchmark datasets.

\textbf{Representative Measurements:} The real world is a complex place for robots. To test the ability of distributed C-SLAM methods to operate in such conditions, we need the measurements in our datasets to reflect this complexity (e.g. measurements with non-parametric/non-stationary noise, measurements corrupted by outliers). Due to the intractability of simulating such conditions, the only way to generate such measurements will be to derive them using front-end algorithms parsing real-world data.

\textbf{Loop-Closures:} The key advantage that C-SLAM algorithms provide is to bound drift using loop-closures both within a robot's trajectory and between robots. We, therefore, will need to derive our benchmark datasets from data that contains plentiful loop-closure opportunities.

\textbf{Long Traversals:} Over short distances $O(100m)$, odometry algorithms can maintain state estimates that are sufficiently accurate for downstream tasks (e.g. planning)~\cite{lidar_odom_survey_lee_2024}. Therefore, for the effects of C-SLAM (i.e. inclusion of loop-closures) to actually be observable, and in turn for our benchmarks to be useful, we will need to derive our benchmark datasets from data of long traversals (i.e. $>1km$).

\textbf{Accurate Reference Solutions:} With our datasets supporting observing the effects from different distributed C-SLAM back-ends, we will also need them to support metric computation to quantitatively evaluate these differences. To support this, we will need our benchmarks to include high-quality reference solutions (often called ``groundtruth'') as well as reference outlier measurement classifications.

\textbf{Temporal Information:} In addition to metric accuracy, a key performance measure for back-ends is their ability to compute solutions to problems in real-time. To support analysis of this performance, we will further require benchmark datasets that maintain accurate temporal information.

\textbf{Convenient:} For benchmark datasets to be useful, the community should want to use them. To facilitate this, the benchmark datasets should be convenient and easy to use so that researchers adopt them to test and evaluate their work. 

It is only by meeting all of these requirements that our benchmark datasets will properly support researchers to test and evaluate novel C-SLAM back-end algorithms.

%% file: sections/2-related_work.tex
\section{Related Work}\label{sec:related-work}

Due to the lack of standard benchmark datasets, prior works have primarily taken three general approaches to evaluating the performance of proposed algorithms: 1) simulated multi-robot datasets~\cite{ddfsam2_cunningham_2013, dgs_choudhary_2017, geod_cristofalo_2020, dc2pgo_tian_2021, asapp_tian_2020, idfgo_matsuka_2023, maj_min_journal_fan_2024, robot_web_journal_murai_2024,  mesa_mcgann_2024, imesa_mcgann_2024, hyperion_hug_2024}, 2) partitioned single-robot benchmark datasets~\cite{admmslam_choudhary_2015, geod_cristofalo_2020,  dc2pgo_tian_2021, asapp_tian_2020,  idfgo_matsuka_2023, maj_min_journal_fan_2024, mesa_mcgann_2024}, and 3) closed-source real-world data~\cite{dgs_choudhary_2017, geod_cristofalo_2020}. While each can support the efficacy of an algorithm, even together they don't permit accurate evaluations and comparisons.

Simulated datasets are an excellent tool to evaluate C-SLAM back-end algorithms. They can be used to evaluate algorithms up to and beyond the bounds of current robotic hardware limitations (e.g. runtime, physical mobility, and team size). Simulated datasets, however, will never be capable of capturing the nuances of real-world data.

Partitioned single-robot benchmark datasets, when based on real-world data, bring increased realism. However, simply partitioning measurements from a single trial fails to accurately represent the topological structure, viewpoint variance, or measurement distribution that we expect from a multi-robot team operating in tandem and communicating to derive inter-robot measurements. Additionally, these benchmarks lack outlier measurements and, in their current formats (\texttt{g2o}~\cite{g2o_kummerle_2011} and \texttt{toro}~\cite{toro_grisetti_2009}), lack temporal information.

A small number of works that develop distributed C-SLAM back-ends have evaluated on real-world data from multi-robot teams~\cite{dgs_choudhary_2017, geod_cristofalo_2020}. Both test on closed-source multi-robot data. While this evaluation is great evidence for the performance of these methods, the closed-source data does not provide a means to reproduce results nor help current researchers in the development of new algorithms.

Open-source multi-robot sensor data does exist and is being released with increased frequency~\cite{graco_zhu_2023, kimeramulti_lessons_tian_2023, s3e_feng_2024, diterpp_kim_2024, coped_zhou_2024, cumulti_albin_2025}. This data, however, has not seen widespread adoption, as it requires researchers to implement their own distributed C-SLAM front-end, which can be a time-intensive and complex task. Additionally, if all researchers implement their own C-SLAM front-ends to utilize this data, then differences in these front-ends are likely to dominate performance differences and obfuscate back-end comparisons between works.

Though they have seen only limited use, there are two datasets that are close to achieving the requirements for our distributed C-SLAM benchmark datasets.

First is the University of Toronto Institute for Aerospace Studies Multi-Robot Cooperative Localization and Mapping benchmark~\cite{utias_leung_2011}. While this benchmark can be an excellent resource for researchers, its artificial construction and controlled lab environment mean that it is not representative of the data we expect from deployed multi-robot systems.

Second is the Nebula multi-robot benchmark~\cite{lamp2_chang_2022}. This benchmark consists of four datasets from multi-robot deployments in subterranean tunnel environments for the DARPA Sub-T Challenge. These datasets consist of 3D pose-graphs that were generated at operation time by a centralized collaborative SLAM system (LAMP 2.0~\cite{lamp2_chang_2022}). The Nebula datasets provide an excellent representation of real-world multi-robot data and include outlier measurements, temporal information, good intra/inter-robot loop-closures, and decent reference solutions from LiDAR scan matching to a survey-grade map. However, loop-closures are computed in a centralized manner, making them unrealistic to what we expect from distributed C-SLAM systems, and as it consists of only four datasets in subterranean environments, this benchmark is limited in its ability to test a back-end's generalization.

With only a single, limited benchmark that achieves the requirements outlined in Sec.~\ref{sec:benchmark_reqs}, we are motivated to develop additional benchmark datasets to support the community and facilitate future research.

%% file: sections/3-methodology.tex
\section{Methodology}\label{sec:methodolody}

To develop a suite of distributed C-SLAM benchmark datasets, we first identify open-source real-world data that supports our benchmark requirements (Sec.~\ref{sec:methodology:data_selection}). We then define a baseline C-SLAM front-end used to derive the measurements that make up the benchmarks (Sec.~\ref{sec:methodology:front_end}) and a communication model required for deriving inter-robot measurements (Sec.~\ref{sec:methodology:comm_model}). We finally derive the reference solution, noise models, and measurement classifications required to complete these benchmark datasets (Sec.~\ref{sec:adtl_details}).

\subsection{Data Selection}\label{sec:methodology:data_selection}

To develop realistic benchmark datasets, we must begin with high-quality real-world data. We specifically focus on LiDAR data, as LiDAR-based SLAM systems are currently preferred in research and industry. Therefore, LiDAR-based benchmarks will be representative of real-world deployments for large sections of the robotics community. 

The best real-world LiDAR data would be that gathered by actual multi-robot teams like the recent releases discussed above~\cite{kimeramulti_lessons_tian_2023, graco_zhu_2023, diterpp_kim_2024, s3e_feng_2024, coped_zhou_2024}. These datasets, however, all fail to support our requirements, as each fails to provide at least one of the following -- an accurate reference solution, long trajectories, or opportunities for loop-closures.

Despite the lack of viable multi-robot data, there is another path to derive our benchmark datasets. We can use multiple independent trials of single-robot data gathered at different times in the same environment to simulate a multi-robot data sequence by temporally aligning the data. By using independent trials, we achieve realistic multi-robot viewpoint variance and topological structure. Additionally, by temporally synchronizing the data, we can simulate communication and, in turn, derive realistic inter-robot measurements.

This approach is made possible by our focus on LiDAR data, as it is invariant to visual differences, meaning that even data from different times will be consistent. While there may be physical differences in the environment when data is collected, these changes are not unrepresentative of dynamic objects moving during multi-robot operations.

Specifically, we synchronize multiple trials into a multi-robot sequence by selecting one to act as the anchor time $t^*$. For all other trials, we sample a relative start time $\Delta_i \sim \Nc(0s, 40s)$ and temporally transform its data to act as though the robot had started operation at $t_i$ where:
\begin{equation}\label{eq:temporal_synchronization}
    t_i = t^* + \Delta_i
\end{equation}

\begin{remark}[Synchronization Side-Effects]
    While we synchronize single-robot trials to the same general operation time, we do not synchronize their measurements nor the period in which they are active. Each robot's sensors gather data at slightly different rates and offsets. Additionally, since each trial is a different length, not all robots may be active at once. We consider all robots to be ``off'' until the timestamp of their first measurement. Inversely, after a robot's final local measurement, we assume that the robot stays active but remains stationary such that it can continue to communicate and even derive new inter-robot measurements. 
\end{remark}

For synchronization, however, we still require single-robot data that supports our benchmark requirements. For this, we selected the Multi-Campus Dataset (MCD)~\cite{mcd_nguyen_2024} and the CU-Multi Dataset~\cite{cumulti_albin_2025}. Together the datasets consist of multiple trials collected by four different platforms in five unique university campus environments. The CU-Multi dataset was even designed intentionally for multi-robot applications, making an argument for temporal synchronization similar to ours. For additional details on these datasets, please see the original publications. To handle these datasets and ensure proper calibration, we utilize \texttt{evalio}~\cite{evalio_arxiv_potokar_2025}.

\subsection{C-SLAM Front-End}\label{sec:methodology:front_end}
To generate benchmarks from synchronized data, we next implement a representative C-SLAM front-end algorithm consisting of an odometry algorithm, a keyframe selection method, a loop-closure detection module, and a loop-closure computation method. For each, we use existing methods representative of the state of the art. 

\begin{figure*}[t]
    \centering
    \begin{subfigure}{2.33in}
        \centering
        \includegraphics[width=1.5in]{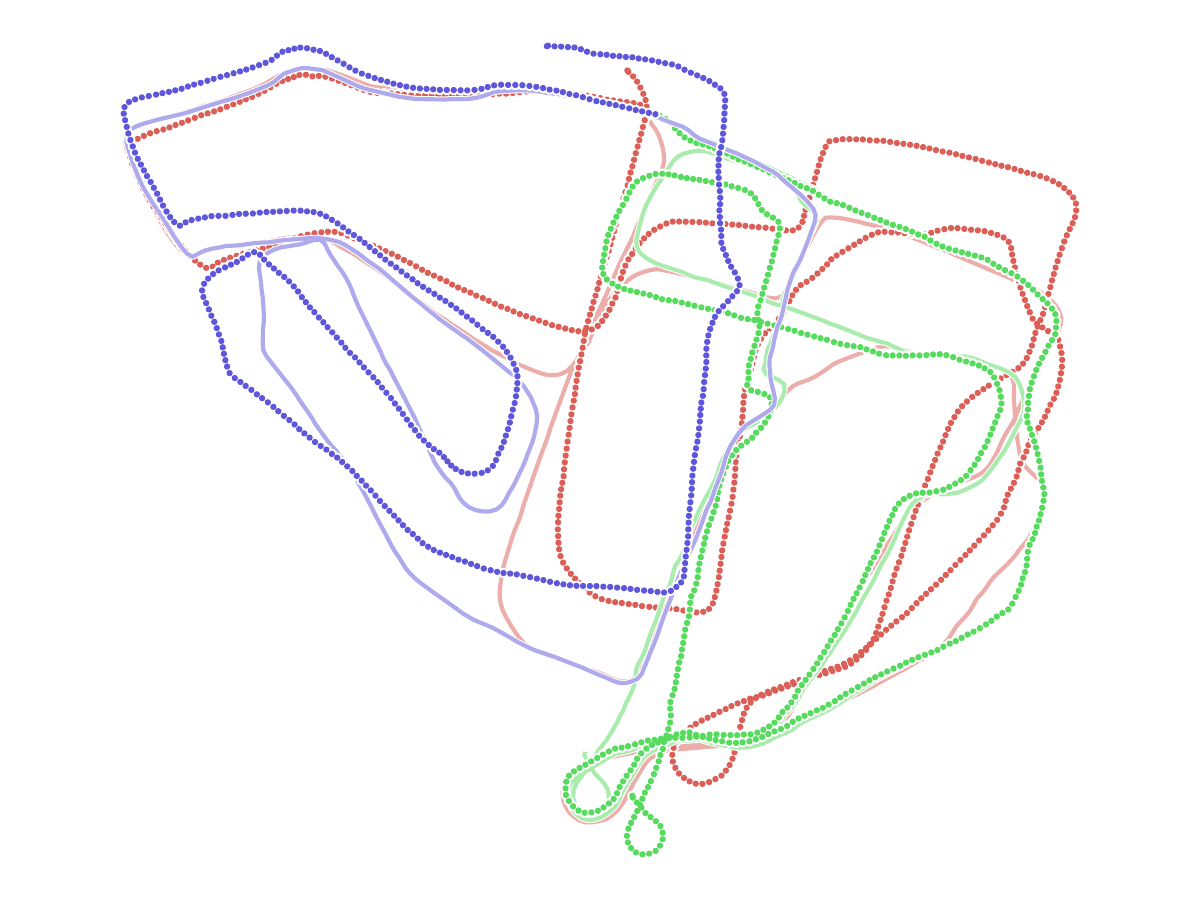}
        \caption{Odometry + Keyframes}
        \label{fig:front_end_results:odometry_keyframes}
    \end{subfigure}%
    \begin{subfigure}{2.33in}
        \centering
        \includegraphics[width=1.5in]{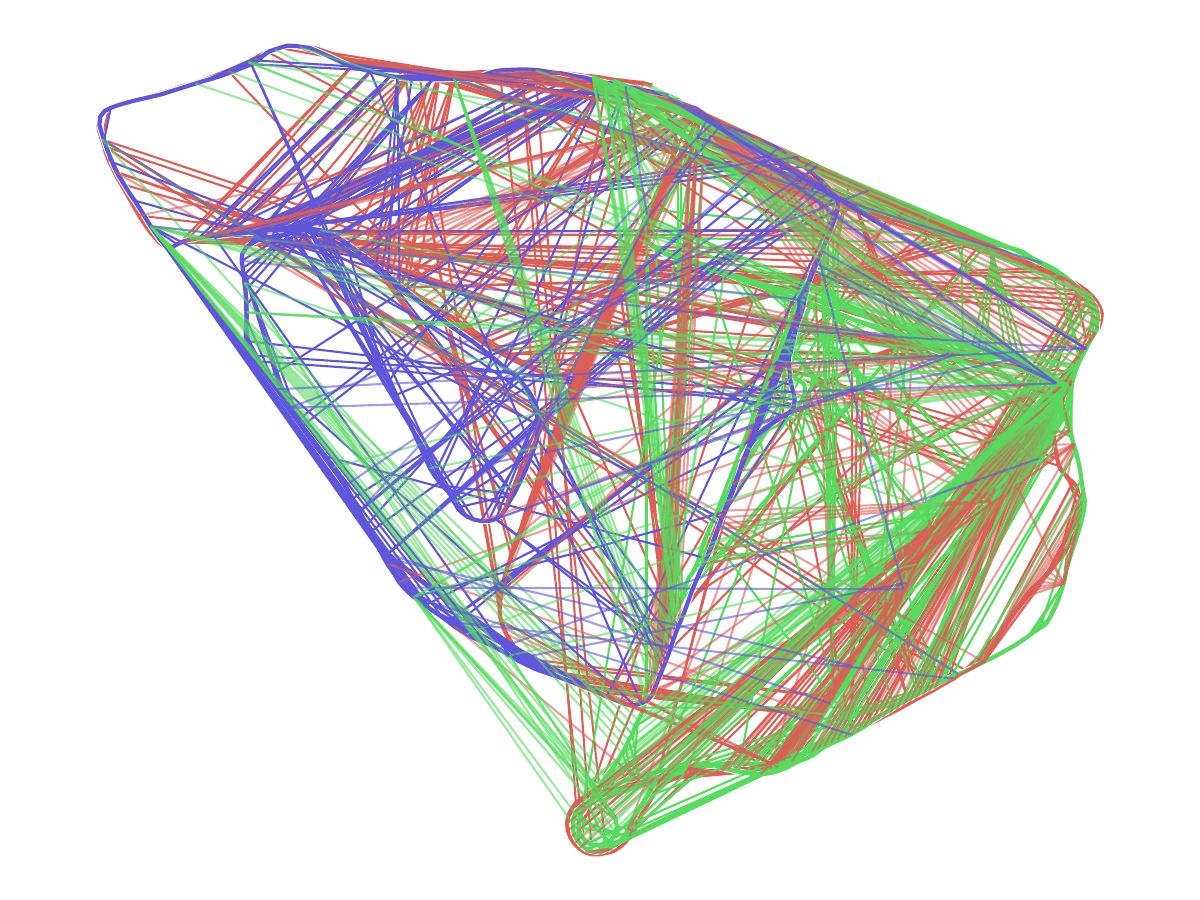}
        \caption{Loop-Closure Detections}
        \label{fig:front_end_results:loop_closure_detections}
    \end{subfigure}%
    \begin{subfigure}{2.33in}
        \centering
        \includegraphics[width=1.5in]{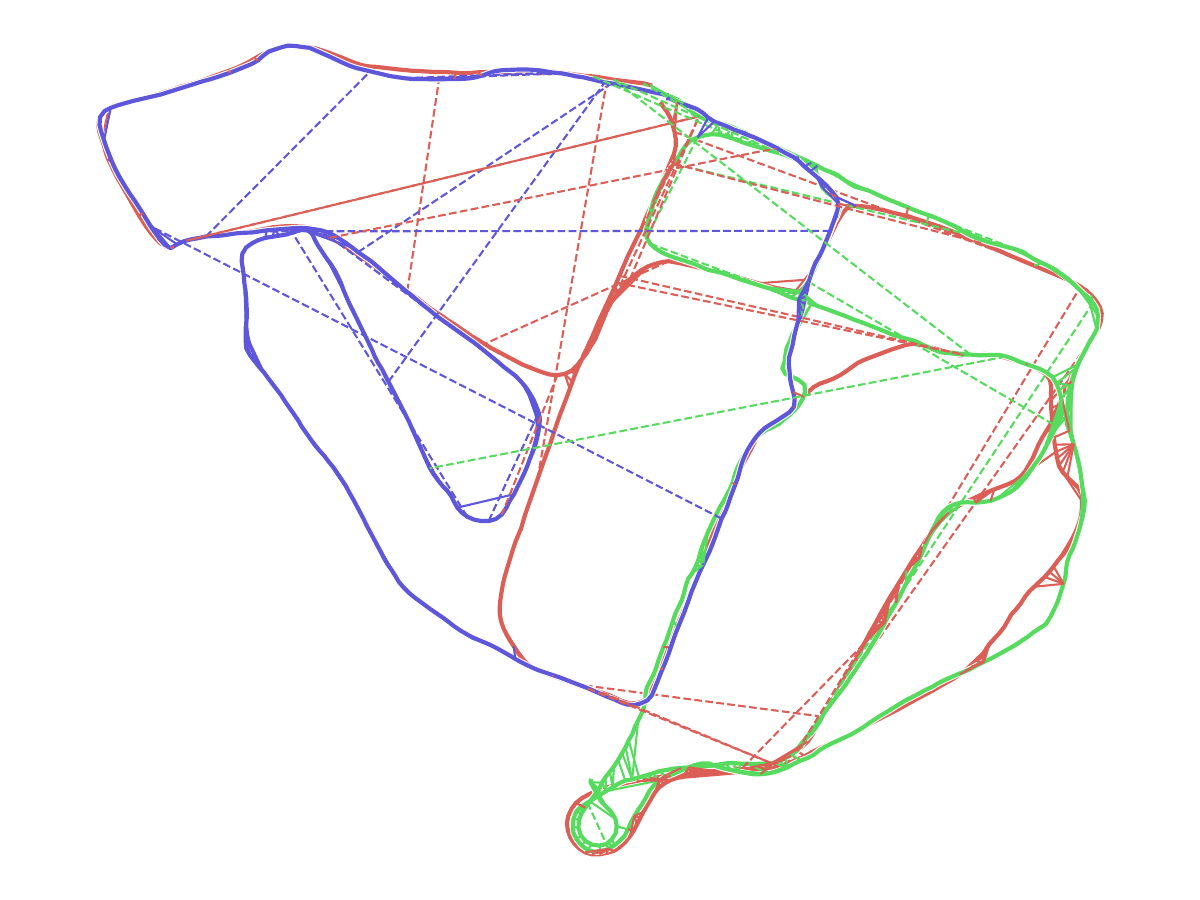}
        \caption{Loop-Closure Measurements}
        \label{fig:front_end_results:loop_closure_measurements}
    \end{subfigure}%
    \vspace{-4pt}
    \caption{Example results from each step in our distributed C-SLAM front-end on the \texttt{kth\_r3\_00} dataset. \ref{fig:front_end_results:odometry_keyframes} depicts the odometry estimated by LOAM in solid colors, the reference trajectories in opaque colors, and the selected keyframes as dots on the odometry estimate, with each color representing one of the 3 robots. \ref{fig:front_end_results:loop_closure_detections} depicts all potential loop-closures detected by ScanContext, with detections colored based on the robot that generated the detection. \ref{fig:front_end_results:loop_closure_measurements} depicts the final set of measurements computed by KISS-Matcher, with inliers drawn with solid lines and outliers drawn with dashed lines.}
    \label{fig:front_end_results}
    \vspace{-0.8cm}
\end{figure*}

\subsubsection{Odometry}
Odometry measurements, observed at the rate of LiDAR scans, are derived using LiDAR Odometry and Mapping (LOAM) -- a standard baseline for LiDAR-only odometry~\cite{zhang_loam_2014}. We use an open-source implementation of LOAM provided through \texttt{evalio}.\footnote{\url{https://github.com/contagon/evalio}}

\subsubsection{Keyframe Selection}
From this high-rate odometry, we sub-sample keyframes -- a standard practice in SLAM pipelines to reduce the problem size. We use a distance-based keyframe selection metric to sample keyframes when the odometry measurements indicate that the robot has traveled more than a threshold distance $d_{kf} = 2m$. Odometry between keyframes is computed directly by integrating the high-rate odometry measurements.

\subsubsection{Intra-Robot Loop-Closure Detection}
For each keyframe, we perform intra-robot loop-closure detection using ScanContext~\cite{kim_scancontext_2018}. A ScanContext descriptor and RingKey are computed for the keyframe scan and compared against a database of descriptors from all prior keyframes using the ``Two-Phase Search'' method described in the original work. We use an open-source implementation of the ScanContext descriptor and database.\footnote{\url{https://github.com/DanMcGann/scan_context}}

\subsubsection{Inter-Robot Loop-Closure Detection}
To detect and compute inter-robot measurements, robots communicate raw LiDAR scans (for communication details, see Sec.~\ref{sec:methodology:comm_model}). When a scan is received from a teammate, we again use ScanContext and perform the ``Two-Phase Search'' against the database of local keyframe descriptors. We then store the descriptor in a unique database for scans received from the teammate. When a local keyframe is selected, it is also compared against the databases for all teammates.

\subsubsection{Loop-Closure Computation}
For each detected loop-closure (both intra-robot and inter-robot), when detected, we attempt to compute a loop-closure measurement using KISS-Matcher~\cite{lim_kissmatcher_2024}. We use the open-source implementation of KISS-Matcher provided by the original authors.\footnote{\url{https://github.com/MIT-SPARK/KISS-Matcher}} This algorithm does attempt to reject outlier measurements, but as expected, it does compute spurious measurements both due to poor detections as well as limitations in the LiDAR scan alignment, resulting in realistic outliers that we want for our benchmark datasets.

\begin{remark}[Dewarping]
    To decouple the results of the odometry module and the loop-closure detection/measurement modules, we dewarp LiDAR scans using the reference solution before passing each scan to the detection/measurement modules.
\end{remark}

This front-end produces the measurements that form the basis of our benchmark datasets. An example of the results from this front-end can be seen in Fig.~\ref{fig:front_end_results}. For completeness, we also include all hyperparameters used for the front-end component algorithms alongside the online dataset release.

\subsection{Communication Model}\label{sec:methodology:comm_model}
To compute a realistic set of inter-robot loop-closure measurements using the pipeline above, we need to accurately model communication between robots. In this section, we derive a model from real-world data that can be used to simulate inter-robot communications for this task.

\subsubsection{Communication Data Analysis}
Open-source data on inter-robot communication during a multi-robot deployment is very limited. Recent data collected by Lajoie et al., however, provides us with an opportunity to analyze the characteristics of inter-robot communications~\cite{mars_analog_lajoie_2025}. 

The communication data was collected during a three-robot experiment in a planetary analog environment. As robots traversed, they communicated using an ad-hoc Wi-Fi network. The authors utilized network profiling tools to record the throughput and latency between all robots over periods of $1s$. In their analysis, Lajoie et al. note that the dominant indicator for network behavior is inter-robot distance, which we utilize below to summarize the data for each channel shown below in Fig.~\ref{fig:mars_analog_comms}.

\begin{figure}[H]
    \centering
    \includegraphics[width=1.68in]{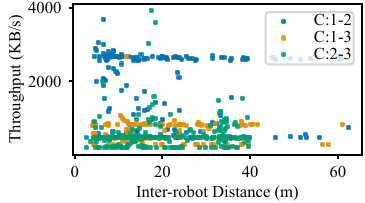}%
    \includegraphics[width=1.68in]{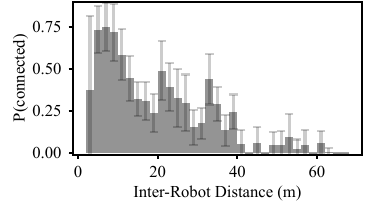}
    \caption{Summary of Wi-Fi communication data recorded by Lajoie et al.~\cite{mars_analog_lajoie_2025}, including per-channel throughput (left) and probability of inter-robot connectivity (right) against inter-robot distance. In the connectivity plots, we additionally plot $99\%$ confidence intervals, though these are overly optimistic as the data is correlated.}
    \label{fig:mars_analog_comms}
\end{figure}

From these figures, we can draw some conclusions about the communication network. Firstly, we can observe that bandwidth is uncorrelated with inter-robot distance; rather, when robots are able to connect, they appear to utilize all the available bandwidth. However, we can see that the bandwidth is not equally shared; we hypothesize that this is a side effect of running multiple instances of network profilers as each attempts to flood the network. When used for actual inter-robot communication, we predict that better network sharing is possible. Second, while bandwidth is not correlated with inter-robot distance, connectivity is correlated. Notably, connectivity between robots is probable at close range and unlikely at larger inter-robot distances.

\subsubsection{Generic Communication Model} 
Using the analysis of this data, we defined a generic communication model that can be used to simulate inter-robot communication.

We assume that all robots engage in a single peer-to-peer communication at a time (to minimize interference) on a shared channel (e.g. Wi-Fi, UWB, acoustic) that has a maximum bandwidth $\Bc$. Simultaneous communications will cause interference and require bandwidth sharing, which we assume to be allocated equally. However, this interference will be spatially localized due to limited communication ranges. Specifically for a communication between robots $a$ and $b$, there will be interference if there exists an actively communicating robot within a distance $d_{intf}$ meters of either $a$ or $b$. We define the total number of communications occurring around $a$ and $b$ at timestep $t$ as $\Ic_{ab}(t, \Theta^*)$, which uses the reference solution $\Theta^*$ to compute inter-robot distances. From this we can define the available bandwidth between robots $a$ and $b$ as $\Bc / \Ic_{ab}(t, \Theta^*)$.

Unlike bandwidth, according to the analysis of Fig.~\ref{fig:mars_analog_comms}, connectivity between agents is correlated with the inter-robot distance. We assume that this correlation is modeled by a function $\phi$ that, given an inter-robot distance $d_{ab}(t, \Theta^*)$ computed from the reference solution $\Theta^*$, outputs the probability that the robots $a$ and $b$ can successfully communicate at this timestep $t$. This allows us to sample the connectivity $\Cc(t)$ between two robots as $\Cc_{ab}(t, \Theta^*) \sim Bern(\phi(d_{ab}(t, \Theta^*)))$.

Together these models for bandwidth and connectivity allow us to simulate the throughput $\Tc$ between robots $a$ and $b$ at a time $t$ over a small timestep $\delta$ (during which we assume that all network traits are constant) as:
\begin{equation}
    \label{eq:throughput}
    \Tc_{ab}(t) = \underbracket[0.5pt][2pt]{\left(\Bc / \Ic_{ab}(t, \Theta^*)\right)}_{\mathrm{Bandwidth}} \cdot \underbracket[0.5pt][2pt]{\Cc_{ab}(t, \Theta^*)}_{\mathrm{Connectivity}} \cdot~\delta
\end{equation}

Putting all of this together, we can define a simple process to simulate inter-robot communications as outlined in Alg.~\ref{alg:communication_simulation}.

\begin{remark}[Communication Initialization] \label{remark:comm_init}
To utilize this model, we need a method to initiate communication between robots. We expect that transmissions will require multiple timesteps $\delta$ of connectivity. To help ensure this, we define an initialization threshold $d_{init}$ based on $\phi$ such that robots initialize communications only when they are close enough to have relatively reliable throughput to complete the communication. An attempt to initialize a new communication is successful if the two robots connectivity $\Cc_{ab}$ is successfully sampled according to the model above.
\end{remark}

\begin{remark}[Communication Timeouts]
We also define a communication timeout $T=2s$ to terminate stale connections. This timeout is both representative of typical network protocols and also prevents our simulation from becoming deadlocked if agents travel too far apart while attempting to communicate.
\end{remark}

\begin{algorithm}
\scriptsize 
\captionsetup{font=scriptsize} 
\caption{Simulate Communication}\label{alg:communication_simulation}
    \begin{algorithmic}[1]
    \State \textbf{In: start/end time $t_s/t_e$, step $\delta$, robots $R$, ref. solution $\Theta^*$}
    \State $C \gets \{\}$ \Comment{Active Communications}
    \State $A \gets   R $ \Comment{Available Robots}
    \For{$t$ from $t_s$ to $t_e$ by $\delta$}
        \State $C, A \gets \mathrm{InitializeNewComms}(t, A, \Theta^*)$ \Comment{Remark~\ref{remark:comm_init}}
        \For{$c\in C$}
            \State $c\gets$ UpdateActiveComm($c, t, \Theta^*$) \Comment{Sample+Integrate Eq.~\ref{eq:throughput}}
            \State $A\gets$ CheckTermination($c, t$) \Comment{Complete/Fail/Maintain}
        \EndFor
    \EndFor
    \end{algorithmic}
\end{algorithm}

\subsubsection{Concrete Models}
From this general model structure, we define two concrete instances, defining the parameters of the model based on expectations for common hardware. First is a Wi-Fi-based model derived from the communication data by Lajoie et al.~\cite{mars_analog_lajoie_2025}. The closed-form equation for $\phi(d) = \min\left(P_{max}, \alpha\left(\beta^{d/r_{max}} - \beta\right)\right)$ was derived from manually fitting a curve to the connectivity data in Fig.~\ref{fig:mars_analog_comms}.

\begin{figure}[h]
    \centering
    \begin{tabular}{@{}c@{}c@{}}
        \begin{minipage}[b][1.25in][t]{1.25in}
        \renewcommand{\arraystretch}{1.1}
        \setlength{\tabcolsep}{2pt} 
        \begin{tabular}{@{}|c|c|c|@{}}
            \hline
            \multicolumn{3}{|c|}{Model Hyperparameters}    \\ \hline
            Parameter      & Wi-Fi      & Pro-Radio         \\ \hline
            $\Bc$ (KB/s)   & 2000       & 1000              \\ \hline
            $d_{init}$ (m) & 30         & 150               \\ \hline
            $d_{intf}$ (m) & 40         & 150               \\ \hline
            $P_{max}$      & 0.7        & 0.8               \\ \hline
            $\alpha$       & 1.1        & 1.8               \\ \hline
            $\beta$        & 0.1        & 0.3               \\ \hline
            $r_{max}$ (m)  & 70         & 200               \\ \hline
        \end{tabular}
        \end{minipage}&
        \includegraphics[width=2in]{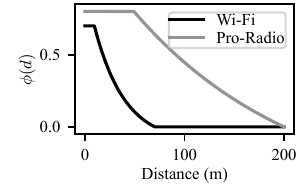}
    \end{tabular}
    \vspace{-0.25cm}
    \caption{Model hyperparameters (left) and connectivity functions $\phi(d)$ (right) for the Wi-Fi and Pro-Radio communication models.}
    \label{fig:comm_models}
\end{figure}

Based on the Wi-Fi model, we also define a model based on ``Professional'' radio equipment (Pro-Radio). Such hardware is expected to have increased range but lower bandwidth due to the increased power requirements. The remaining hyperparameters are scaled from those of the Wi-Fi model to maintain a similar structure. Both the Wi-Fi and Pro-Radio models are summarized in Fig.~\ref{fig:comm_models}.

\subsubsection{Interactions with Front-End}
To utilize these concrete models to simulate inter-robot communication for the purpose of supporting the front-end described above, there are two additional details to specify -- the selection of LiDAR scans to send when a communication is initialized and the amount of data required to transmit that scan.

LiDAR scan selection is performed naively. When a communication is initialized, the transmitting robot randomly selects (uniformly) a keyframe scan that has not yet been successfully sent to the receiving robot.

We compute the data transmission size of a LiDAR scan assuming that agents only need to send $[x,y,z]$ data represented by 32-bit floating-point numbers, that the packet header sizes are negligible relative to the size of the scan, and that the data is first compressed using \texttt{xz}.\footnote{\url{https://github.com/tukaani-project/xz}} We assume that the compression factor provided by \texttt{xz} is modeled by a normal distribution such that the compressed size $\mathfrak{s}$ can be computed from the original size $S$ as $\mathfrak{s} = f \cdot S$ and the compression factor $f$ is sampled from $\Nc(\mu_{xz} = 0.653, \sigma_{xz} = 0.04)$. Where we empirically computed this distribution using a random sample of 20 LiDAR scans from the KTH Day 6 trial of the Multi-Campus Dataset~\cite{mcd_nguyen_2024}.

\begin{remark}[Communication Model Limitations] \label{remark:communication_model_limitations}
While designed for accuracy, our inter-robot communication model has non-trivial limitations. The one-active-communication assumption, equal-bandwidth-sharing assumption, and disregard of packet latency may not be perfectly reflective of actual operations. However, more impactfully, we know that our distance-based connectivity model will not reflect reality as it ignores physical interference (e.g. buildings, vegetation, walls, etc.). While our connectivity models $\phi$ are based on data collected in the presence of physical interference, lack of explicit modeling limits our model's accuracy.
\end{remark}

\subsection{Additional Details}\label{sec:adtl_details}
For completeness, in this section we provide a few final details on the dataset derivation process for COSMO-Bench.

\subsubsection{Prior}\label{sec:adtl_details:prior}
For all datasets, we assume that all robots have an accurate initial estimate in the global reference frame, which is taken from the reference solution. Since this estimate is quite accurate, we manually define a noise model with a rotational standard deviation of $\sigma_r=1e^{-4}~rad$ and a translational standard deviation of $\sigma_t=1e^{-3}~m$. 

\subsubsection{Reference Solution}\label{sec:adtl_details:reference_solution}
Deriving a reference solution for our benchmarks is simple due to our data selection. Both the MCD and CU-Multi datasets have accurate reference solutions provided by alignment to a survey-grade map and a LiDAR-Inertial + RTK-GPS + Digital Elevation Model SLAM algorithm, respectively~\cite{mcd_nguyen_2024, cumulti_albin_2025}. To derive a reference solution for our benchmarks, we simply extract the poses from the provided references for each keyframe.

\subsubsection{Measurement Noise Models}\label{sec:adtl_details:measurement_noise_models}
This reference solution also permits us to derive measurement noise models. For each data collection platform, we select a single trial and run it through the front-end above without any robot team. Using the computed measurements $m \in SE(3)$ and their respective ``true'' measurements $m^*$ (derived from the reference solution), we compute a sample covariance $Q$ for both the odometry measurements and loop-closure measurements. Since both are relative pose measurements, covariance is computed in the tangent space $\mathfrak{se}(3)$ according to:
\newcommand{\outerprod}[2]{\left\lvert#1\right\rangle \left\langle#2
\right\rvert}
\begin{equation}\label{eq:relative_pose_covariance}
    Q = \sum_{m \in M} r^{}_m r^\top_m
\end{equation}
where $r_m=\logmap{m^{-1}\circ m^*}$, which is defined using standard Lie Group notation~\cite{microlie_sola_2021}. One confounding factor for computing this covariance is that it assumes none of the measurements are outliers. For loop-closure measurements that contain outliers, we compute the sample covariance on only ``good'' measurements, defined as having a translational and rotational error (as compared to the reference solution) less than a user-supplied threshold (e.g. $0.5~m$ and $0.05~rad$).

\begin{remark}[Analytical Noise Models]
    We would prefer to use analytic noise models for each measurement over empirical models. However, the component algorithms we use in our front-end do not provide estimates for measurement noise, as doing so is extremely difficult due to iterative re-association within the algorithms~\cite{icp_cov_censi_2007}.
\end{remark}

\subsubsection{Measurement Classification} \label{sec:adtl_details:measurement_classification}
Finally, we make use of these noise models to define the final aspect of our dataset -- reference outlier classifications. We assume that all measurements are affected by Gaussian noise. This means that the residuals for the measurements will be $\chi^2$ distributed. We can use this fact along with the empirically computed noise models to classify our measurements. Specifically, we define a loop-closure measurement as an outlier if its measurement residual (weighted by the noise model above) exceeds the $95\%$ $\chi^2$ critical value.

%% file: sections/4-cosmobench-datasets.tex
\section{COSMO-Bench}\label{sec:datasets}

Using the process described in Sec.~\ref{sec:methodolody}, we generate a total of 24 multi-robot benchmark datasets that make up COSMO-Bench. More specifically, we define 10 sequences using trials from the Multi-Campus data\footnote{Note: The trial NTU Day 10 from MCD is omitted from any sequences due to issues with its reference solution.} and 2 from the two locations in the CU-Multi data. Due to the fact that the communication model will affect the data passed between teammates and, in turn, which inter-robot measurements are computed, we generate one dataset for each communication model for all sequences, resulting in a total of 24 datasets. For both models, we use $\delta=0.1s$ for communication simulation. For a summary of the COSMO-Bench datasets, see Fig.~\ref{fig:datasets}.

\begin{remark}[Noise Models] 
For all sequences in a common environment, we use a constant empirical noise model derived from a single trial. The trials used for noise models are as follows -- NTU=Night 4, KTH=Day 6, TUHH=Day 2, and CU Boulder Campus=Kittredge Loop Robot 3.
\end{remark}

We make all benchmark datasets publicly available in JSON Robot Log (JRL) format.\footnote{\url{https://github.com/DanMcGann/jrl}} Each dataset contains ordered prior/odometry/loop-closure measurements like familiar formats (e.g. \texttt{g2o}, \texttt{toro}). However, JRL unambiguously separates data per robot, includes temporal information for runtime evaluation, provides reference solutions for algorithm evaluation, and is built on JSON for easy parsing.

%% file: sections/5-neubla-dataset-adaption.tex
\subsection{Nebula Multi-Robot Datasets} \label{sec:nebula_dataset}

For the convenience of the community, we additionally provide the Nebula Multi-Robot Datasets in the same JRL format as our benchmarks (Fig.~\ref{fig:nebula_datasets}).

\begin{figure}[ht]
    \centering
    \include{figs/dataset_figures/nebula_dataset_figures}
    \vspace{-12pt}
    \caption{The Nebula datasets. For each dataset, we plot the reference solution and enumerate metadata including -- The dataset name, involved robots, total duration (MM:SS), total distance traveled, and the number (\#) of measurements plus outlier rate (\%) for both intra-robot (LC) and inter-robot (IRLC) loop-closures (\#, \%).}
    \label{fig:nebula_datasets}
\end{figure}

The Nebula datasets are provided as global factor graphs containing measurements generated by LAMP 2.0. To convert to JRL, we simply sequenced each robot's odometry measurements and distributed the loop-closure measurements. We additionally derive a new noise model for the datasets. Comparing the measurements in these datasets against the reference solution suggests that the recorded noise models are overly optimistic. We derive a new noise model for both odometry and loop-closure measurements empirically (Sec.~\ref{sec:adtl_details:measurement_noise_models}) using the \texttt{tunnel} dataset. We additionally use this model along with the reference solution to classify inlier/outlier measurements (Sec.~\ref{sec:adtl_details:measurement_classification}).

\begin{remark}[The \texttt{urban} Reference Solution]
WARNING: Careful inspection of the map produced by the reference solution of the \texttt{urban} datasets shows misaligned LiDAR scans, indicating that the reference solution provided for this dataset is somewhat inaccurate.
\end{remark}

%% file: figs/dataset_figures/nebula_dataset_figures.tex
\setlength{\fboxsep}{0pt}

\newcommand\nebulainfotable[6]{
\begin{minipage}[b][0.75in][c]{0.85in}
\centering
\begin{tabular}{@{}p{0.28in}|c@{}}
    \textbf{Name}           & \multicolumn{1}{p{0.4in}}{\centering\texttt{\textbf{#1}}} \\ \hline
    Robots                  & #2                   \\ \hline
    Dur.                    & #3                   \\ \hline
    Length                  & #4                   \\ \hline
    LC                      & #5                   \\ \hline
    IRLC                    & #6                   \\
\end{tabular}\end{minipage}}

\renewcommand{\arraystretch}{1.1}
\setlength{\tabcolsep}{2pt} 


\begin{tabular}[c]{@{}c@{}c@{}c@{}c@{}}
\includegraphics[trim={3in 0 3in 0},clip,height=0.75in]{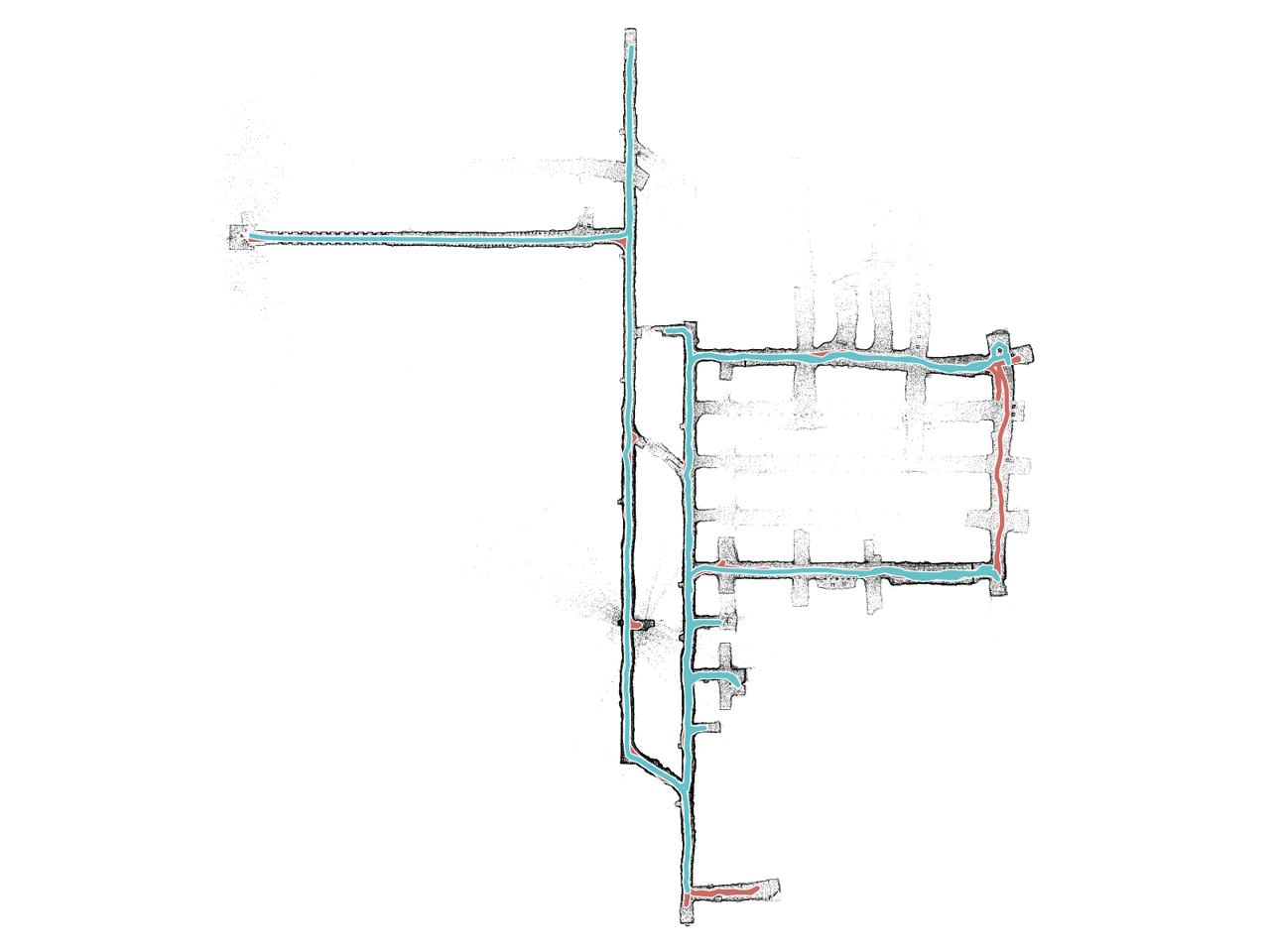} &
\nebulainfotable{tunnel}{d, c}{71:06}{2.59km}{3130(7.0\%)}{2426(8.7\%)} &

\includegraphics[trim={0.8in 0 0.8in 0},clip,height=0.75in]{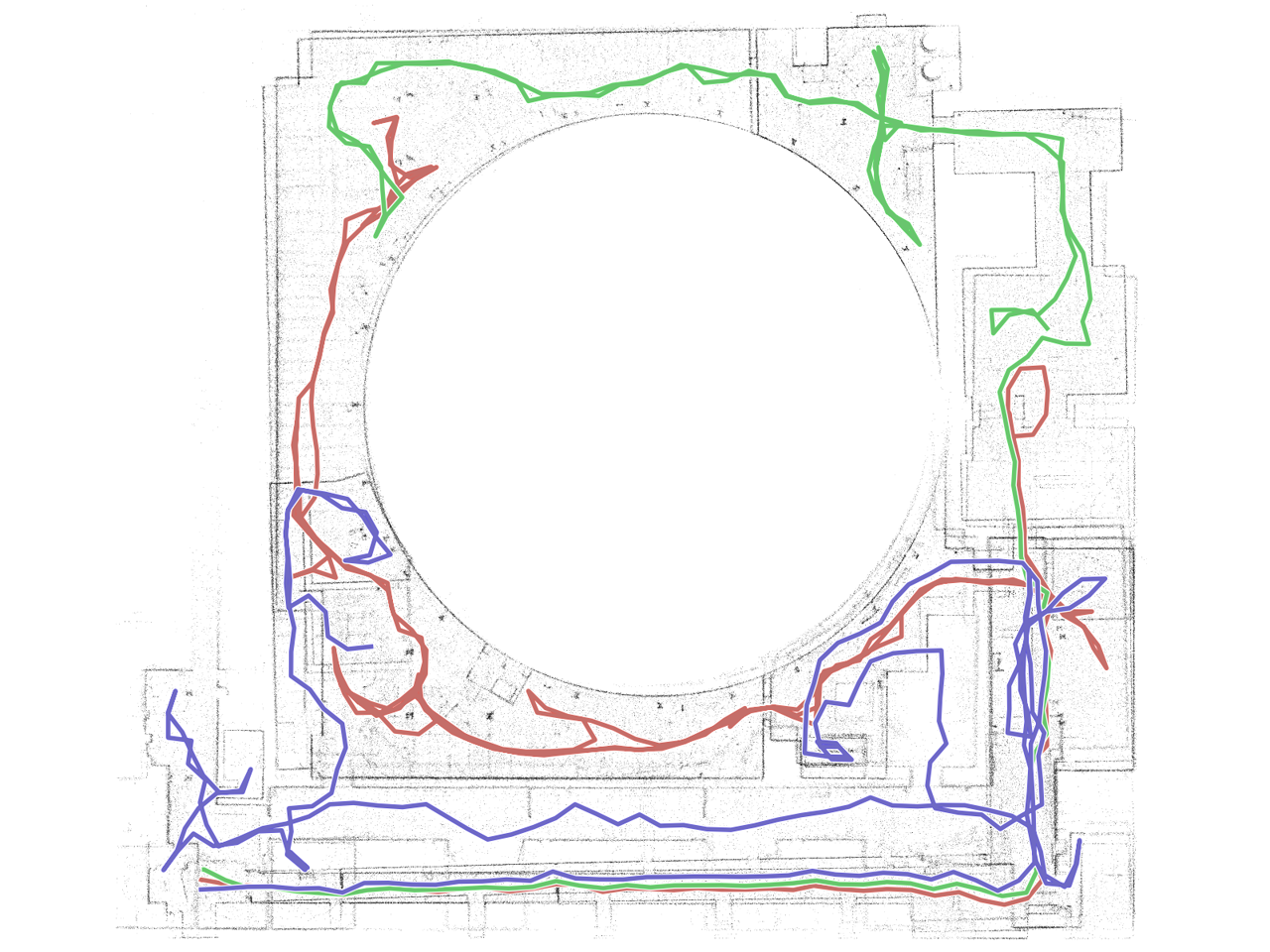} &
\nebulainfotable{urban}{a, d, e}{44:33}{1.55km}{846(21.4\%)}{606(68.5\%)} \\

\includegraphics[trim={3in 0 3in 0},clip,height=0.75in]{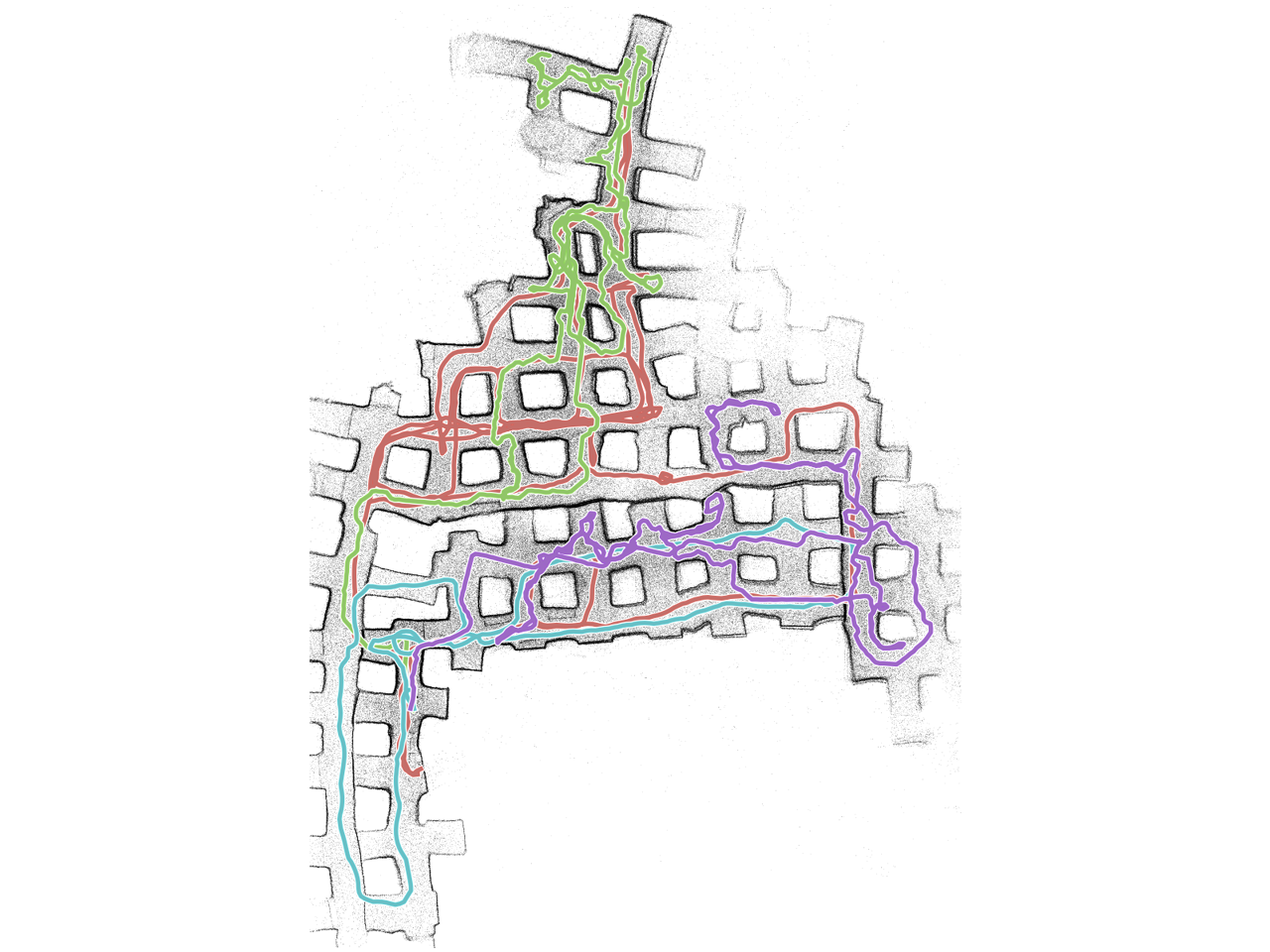} &
\nebulainfotable{\fontsize{6.3}{6.3}\selectfont{ku}}{a, c, d, b}{57:54}{6.01km}{653(5.4\%)}{232(37.1\%)}&

\includegraphics[trim={0.8in 0 0.8in 0},clip,height=0.75in]{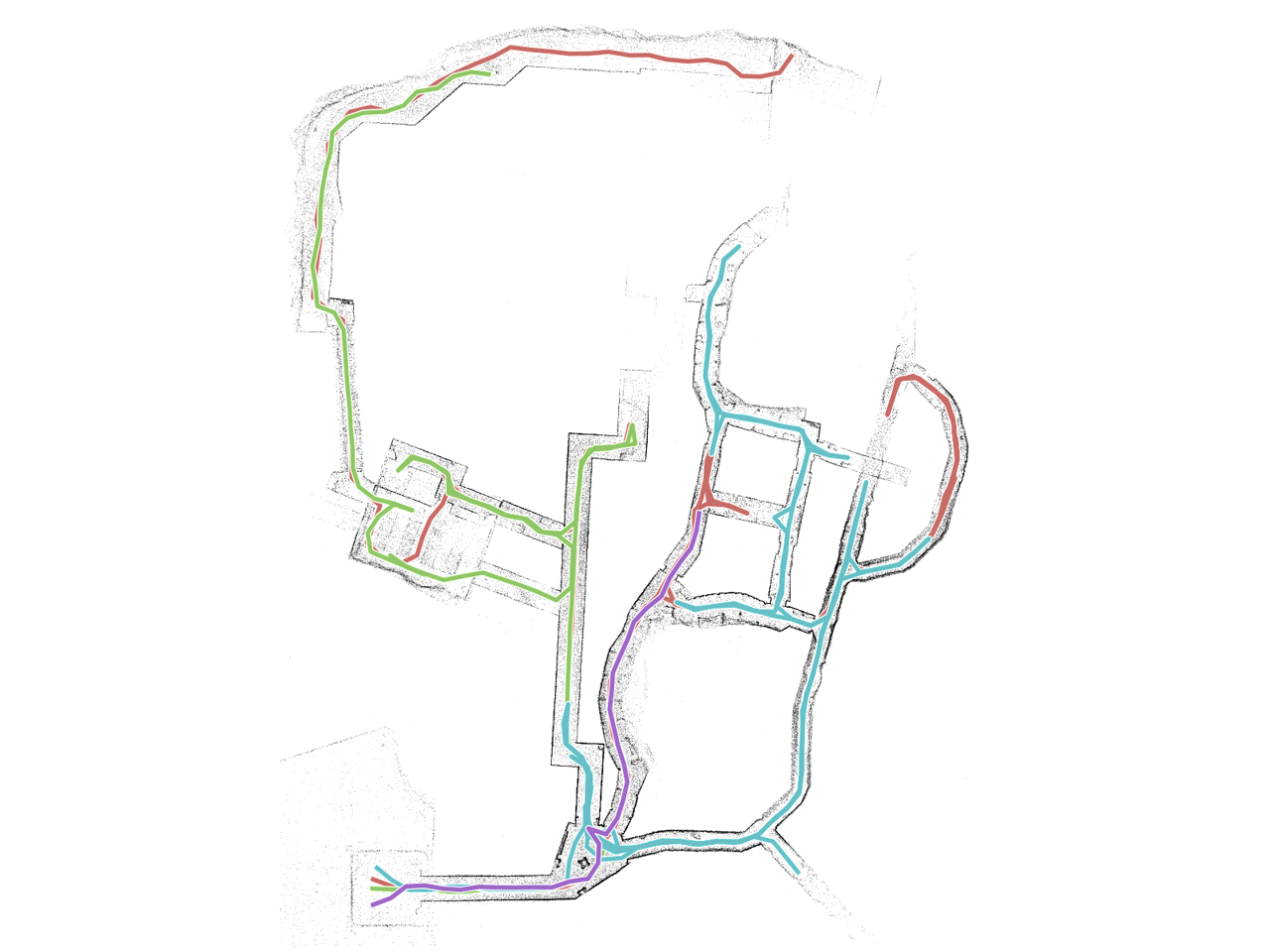} &
\nebulainfotable{finals}{g, h, e, c}{41:12}{1.21km}{662(11.9\%)}{801(29.3\%)}\\

\end{tabular}

%% file: sections/6-conclusion.tex
\section{Conclusion}\label{sec:conclusion}
We hope that COSMO-Bench serves as a useful tool for the community as we collectively work to develop and evaluate the next generation of C-SLAM back-end algorithms.